\documentclass[10pt,twocolumn,letterpaper]{article}
\usepackage[pagenumbers]{cvpr}

\makeatletter
\renewcommand{\paragraph}{%
  \@startsection{paragraph}{4}{\z@}%
    {.35em}%
    {-1em}%
    {\normalfont\normalsize\bfseries}%
}
\makeatother

\definecolor{cvprblue}{rgb}{0.21,0.49,0.74}
\usepackage[pagebackref,breaklinks,colorlinks,allcolors=cvprblue]{hyperref}
\def\confName{CVPR}
\def\confYear{2027}
\usepackage{natbib}
\usepackage[utf8]{inputenc}
\usepackage[T1]{fontenc}
\usepackage{amssymb}
\usepackage{amsmath}
\usepackage{amsbsy}
\usepackage{amsfonts}
\usepackage{amsthm}
\newtheorem{proposition}{Proposition}
\usepackage[nice]{nicefrac}
\usepackage{breqn}
\usepackage{mathtools}
\usepackage{graphicx}
\usepackage{subcaption}
\usepackage{svg}
\usepackage{tikz}
\usepackage{tikz-cd}
\usepackage{pgfplots}
\usepackage{float}
\usepackage{dblfloatfix}
\usepackage{booktabs}
\usepackage{microtype}
\usepackage{multirow}
\usepackage{booktabs}
\usepackage{adjustbox}
\allowdisplaybreaks
\pgfplotsset{compat=1.16}

\usepackage{pifont}
\newcommand{\cmark}{\ding{51}}
\newcommand{\xmark}{\ding{55}}
\title{Color Equivariant Representations Enable Illumination Robust Diffusion Policies}
\author{Yulong Yang$^{1}$ \quad Fan Wu$^{2}$ \quad Christine Allen-Blanchette$^{1}$ \quad Amit Chakraborty$^{3}$\\
$^{1}$Princeton University \quad $^{2}$Arizona State University \quad $^{3}$Siemens Foundational Technologies}
\date{}
\begin{document}
\maketitle

\begin{abstract}
    Image-conditioned diffusion policies inherit sensitivity to lighting through their visual encoders, so differences in the observed illumination can cause their predicted actions to vary even when the underlying task geometry is unchanged. Existing approaches primarily address such sensitivity by using augmentation or observation preprocessing, while the focus of equivariant robot policies has instead centered on achieving spatial equivariance. We introduce CARROT, a new diffusion policy whose vision encoder is equivariant to global luminance shifts and produces invariant spatial conditioning for action generation. This makes the conditional action distribution invariant to these shifts without modifying the denoiser or diffusion objective. We evaluate CARROT under progressively less idealized lighting changes, ranging from a controlled global luminance shift to spatially varying perturbations of illumination intensity, direction, color, and shadow, and finally a measured change in illumination intensity and direction on real robot hardware. In every shifted setting, CARROT achieves the highest task success rate among Diffusion Policy and augmentation, preprocessing, restoration, and point-cloud baselines, while matching the vanilla policy on clean observations. These results establish photometric equivariance as a principled architectural basis for illumination-robust policies.
\end{abstract}
\vspace{-0.15in}
\section{Introduction}
Diffusion policies~\citep{chi2025diffusion,ze20243d} have emerged as a strong approach to visuomotor imitation learning because they represent conditional distributions over action sequences through iterative denoising, allowing them to capture multimodal demonstrations.
In image-conditioned policies, however, the action generator receives the scene only through a learned visual representation. 
Changes in illumination intensity or camera exposure can therefore corrupt the predicted actions even when the scene geometry is unchanged.
This sensitivity is a practical obstacle because demonstrations are often collected under a narrow range of imaging conditions, while deployment environments may be substantially brighter, darker, or differently illuminated~\citep{baek2024unexplored,jin2026robolight,zhang2026rohil}.

On systematic robustness benchmarks, manipulation policies degrade sharply under appearance shifts, including changes in lighting and object color~\citep{pumacay2024colosseum,fei2025libero}.
Controlled real-world studies further show that lighting changes alone reduce task success~\citep{xie2024decomposing}, largely through object errors in object localization~\citep{jin2026robolight}.
Existing approaches mitigate such shifts either by broadening the training distribution or by correcting observations before they reach the policy.
Domain randomization, photometric augmentation, and learned relighting expose the policy to a wider range of appearances during training~\citep{tobin2017domain,hendrycks2021many,zini2022planckian,jin2025physically,zhang2026rohil}, but invariance is only encouraged through the training loss and remains approximate even for the augmented lighting variations.
Observation-processing methods instead suppress or repair appearance through diffusion inversion or low-light enhancement~\citep{wei2018deep,guo2020zero,niu2023vergnet,hu2025stem,sun2026light}, but because they are designed independently of the policy, they can distort well-exposed observations that the policy would otherwise handle correctly.
Point-cloud policies avoid appearance altogether~\citep{ze20243d,chisari2024learning,ze2025generalizable}, but remove color information that some tasks require.
None of these approaches explicitly constrains how the policy representation must change under a known photometric transformation.

Equivariant networks impose this constraint directly.
Each layer is constructed to commute with the action of a known transformation group, so the response of the features to that group is fixed by the architecture rather than learned from examples~\citep{bronstein2021geometric}.
Existing equivariant diffusion policies have primarily targeted spatial transformations, including planar rotations and three-dimensional similarity or rigid transformations~\citep{wang2024equivariant,yang2024equibot,tie2025seed,zhu2025se}, which move scene content rather than alter its appearance.
Color-equivariant vision models have encoded hue, saturation, and luminance transformations as group actions~\citep{lengyel2023color,yang2025lcer,yang2026hypertoroidal,messaoudi2026designing}, but for classification and segmentation rather than control.
For a diffusion policy, which observes the scene only through its visual conditioning, an encoder that is equivariant to such a photometric group and invariant at its output would make the conditional action distribution invariant as well, without modifying the denoiser or diffusion objective.

We introduce \textbf{C}olor-equiv\textbf{A}riant \textbf{R}epresentations for illumination \textbf{R}obust visu\textbf{O}motor con\textbf{T}rol (CARROT), an image-conditioned diffusion policy that builds robustness to global luminance shifts into its vision encoder.
The encoder first lifts each image into a discrete orbit of luminance-transformed copies.
A luminance shift represented by the group therefore cyclically permutes the orbit along an explicit photometric axis rather than deforming the representation arbitrarily.
Group convolutions share filters across this axis~\citep{yang2026hypertoroidal}, and normalization statistics pooled over the complete orbit keep the permutation intact through every layer.
Max pooling over the orbit then removes the luminance coordinate, and an invariant spatial softmax extracts keypoints that retain where task-relevant evidence appears in the image.
Luminance shifts in the group therefore leave the visual conditioning unchanged, while for lighting changes outside the group the change in conditioning is bounded in proportion to how far they lie from the orbit.

We evaluate CARROT under lighting changes that are progressively less idealized, ranging from a controlled global luminance shift to spatially varying illumination in simulation and a measured change in illumination intensity and direction on real robot hardware.
In every shifted setting, including with language-conditioned policies, CARROT achieves the highest average task success rate among vanilla Diffusion Policy and augmentation, preprocessing, restoration, and point-cloud baselines, while matching the vanilla Diffusion Policy and baselines on unperturbed observations.
Together, these results establish photometric equivariance as a principled architectural basis for designing illumination-robust visuomotor policies.
\section{Related Work}
\paragraph{Diffusion Policy under Perceptual Variation.}
Diffusion Policy~\citep{chi2025diffusion} is widely adopted for visuomotor imitation learning, but its image encoder entangles appearance with task structure, so shifts in lighting or object color degrade behavior even when task geometry is fixed~\citep{jin2026robolight,huang2025spatial}.
Point-cloud policies switch the input modality, reducing dependence on RGB appearance~\citep{ze20243d,wang2024rise,chisari2024learning} but discarding chromatic cues that color-dependent instructions require~\citep{jia2026pointmappolicy}.
3D Diffuser Actor~\citep{ke20243d}, GenDP~\citep{wang2024gendp}, and PointMapPolicy~\citep{jia2026pointmappolicy} reintroduce image features into 3D representations, restoring color along with its photometric sensitivity.

Other approaches retain image observations and intervene on the training data, the observation, or the representation.
Domain randomization varies simulated textures and lighting~\citep{tobin2017domain}, generative augmentation synthesizes new scene content~\citep{chen2023genaug}, and physically based relighting re-renders demonstrations~\citep{jin2025physically}.
Observation-processing methods normalize the input through diffusion inversion~\citep{hu2025stem}, task-relevant masking~\citep{mirjalili2025arro}, or low-light enhancement~\citep{niu2023vergnet}.
Disentangled and depth-aware representations separate task structure from nuisance appearance~\citep{batra2025zero,huang2025spatial}.
In each case, robustness is learned from data or delegated to preprocessing rather than built into the policy.
Equivariant diffusion policies do build in structure, but for spatial symmetries such as planar rotations and three-dimensional rigid or similarity transformations~\citep{wang2024equivariant,yang2024equibot,tie2025seed,zhu2025se}, under which the correct action transforms with the scene.
An illumination change leaves the correct action unchanged and instead calls for invariant conditioning.
CARROT builds this invariance into the vision encoder, leaving the denoiser and diffusion objective unchanged.
\paragraph{Equivariant Vision.}
Convolutional networks are equivariant to planar translations~\citep{bronstein2021geometric}, and group convolutions generalize the construction to broader symmetry groups~\citep{kondor2018generalization,cohen2019general}.
The first such networks were equivariant to finite planar rotation groups~\citep{cohen2016group}, and circular harmonic filters extended this to continuous rotations~\citep{worrall2017harmonic}.
Scale equivariance has been obtained through log-polar coordinates~\citep{esteves2017polar}, dilation-based scale spaces~\citep{worrall2019deep}, and Fourier-constrained filters~\citep{sun2023empowering}.
Spherical-harmonic and gauge-equivariant architectures extends the same principle to three-dimensional rotations as well as local coordinate systems~\citep{thomas2018tensor,esteves2018learning,esteves2019equivariant,cohen2019gauge}.

Beyond geometry, vision models are also sensitive to color and illumination changes~\citep{buhrmester2019evaluating,de2021impact}, including in robot learning~\citep{hu2025stem,jin2026robolight}.
Normalization-equivariant networks build equivariance to global intensity scaling and offsets into denoisers~\citep{herbreteau2023normalization}, and affine-invariant networks provide the corresponding invariance for classification and localization~\citep{messaoudi2026designing}.
Color equivariant convolutions share parameters across hue shifts by rotating RGB filters~\citep{lengyel2023color}.
Lifting the input rather than the filters avoids the invalid RGB values of filter rotation and extends equivariance to saturation and luminance~\citep{yang2025lcer}.
Treating these interval-valued channels as translations, however, introduces artifacts.
The hypertoroidal covering removes them by lifting each interval onto a circle on which the shifts act cyclically~\citep{yang2026hypertoroidal}.
These color-equivariant networks were developed primarily for classification, where the readout pools away spatial structure.
CARROT carries the luminance construction into action generation, pairing orbit pooling with a spatial softmax so that the invariant conditioning still localizes task-relevant evidence.
\section{Background}
\paragraph{Diffusion Policy.}
Visuomotor policy imitation learning can be formulated as conditional generation over short action trajectories~\citep{chi2025diffusion,ze20243d}.
At control step $t$, the policy predicts a chunk of future actions $\mathbf{A}_{t}$ from the observation history $\mathbf{O}_{t}$.
Rather than regressing a single action sequence, it models the conditional distribution $P\left(\mathbf{A}_{t}\mid\mathbf{O}_{t}\right)$ through iterative denoising~\citep{ho2020denoising}, which lets it represent multiple valid behaviors in the demonstrations.
Let $\mathbf{X}_{t}$ denote the image history and $\mathbf{S}_{t}$ the nonvisual conditioning, including robot state and task information.
An encoder $E_{\psi}$ maps the images to the conditioning vector $c_{t}=\left[E_{\psi}\left(\mathbf{X}_{t}\right),\mathbf{S}_{t}\right]$, on which the denoiser depends throughout training and sampling.
Visual information reaches the action distribution only through $E_{\psi}\left(\mathbf{X}_{t}\right)$, so a luminance shift can leave task geometry and robot state unchanged while still altering the conditioning and actions.
\paragraph{Equivariance.}
Consider a set $X$ and group $G$, the action of $G$ on $X$ is defined with the function $\varphi:G\times X\rightarrow X$ which satisfies
\begin{enumerate}
    \item Identity: $\forall x\in X$, $\varphi\left(\boldsymbol{1},x\right)=x$ where $\boldsymbol{1}\in G$ is the identity element.
    \item Compatibility: $\forall g,h\in G$ and $\forall x\in X$, $\varphi\left(g,\varphi\left(h, x\right)\right)=\varphi\left(gh, x\right).$
\end{enumerate}
We call a set $X$ equipped with the left-group action of group $G$ a left $G$-set~\citep{humphreys1996course,rotman2012introduction,gallier2020differential}.
Given two left $G$-sets $X$ and $Y$, and the corresponding group actions $\varphi:G\times X\rightarrow X$ and $\phi:G\times Y\rightarrow Y$, a function $f:X\rightarrow Y$ is equivariant~\citep{serre1977linear,fulton2013representation,gallier2020differential} if and only if $\forall x\in X$ and $g\in G$
\begin{equation}
    f\left(\varphi\left(g, x\right)\right) = \phi\left(g, f\left(x\right)\right),
\end{equation}
where invariance is a special case of equivariance when $\phi$ is the identity map~\citep{olver1999classical,bronstein2021geometric}, such that $f\left(\varphi\left(g, x\right)\right) = f\left(x\right)$.
\paragraph{Group Convolutions.}
Equivariance can be enforced in convolutional networks by defining each layer over a symmetry group rather than the pixel grid~\citep{cohen2016group,kondor2018generalization}.
We can write conventional convolution with translational equivariance as
\begin{equation}
    f^{\ell+1}=\left[f^{\ell}*\psi_{i}^{\ell}\right]\left(x\right) = \sum_{y\in\mathbb{Z}^{2}}\sum_{k=1}^{K_{\ell}} f^{\ell}_{k}\left(y\right)\psi_{i,k}^{\ell}\left(x-y\right),
\end{equation}
where $f^{\ell}:\mathbb{Z}^{2}\rightarrow \mathbb{R}^{K_{\ell}}$ denotes the input to layer $\ell$ and $\psi_{i}^{\ell}: \mathbb{Z}^{2}\rightarrow\mathbb{R}^{K_{\ell}}$ denotes the convolutional filters, and $K_{\ell}$ define the filter channel.
This operation can be made equivariant to the actions of group $G$ through group convolution defined as
\begin{equation}
    f^{\ell+1}=\left[f^{\ell}*\psi_{i}^{\ell}\right]\left(g\right) = \sum_{h\in G}\sum_{k=1}^{K_{\ell}} f^{\ell}_{k}\left(h\right)\psi_{i,k}^{\ell}\left(h^{-1}g\right).
\end{equation}
We leverage this property to design a perceptually equivariant vision encoder in the following section.
\section{Method}\label{sec:method}
CARROT combines the diffusion policy and group convolution frameworks above, replacing the standard visual encoder with a luminance-equivariant network while leaving the diffusion denoiser and training objective unchanged.
Rather than asking the encoder to suppress an unfamiliar illumination change, we make that change explicit in the representation.
Lifting the observation onto a discretized luminance group turns a lighting change represented by the group into a permutation of the features rather than an unpredictable deformation, and group convolutions preserve this structure throughout the encoder.
An invariant readout then removes this photometric degree of freedom while preserving the spatial information required for control, so the denoiser receives the same visual conditioning.
\subsection{Equivariant Vision Encoder}\label{subsec:method_encoder}
\begin{figure}[t]
    \centering
    \includegraphics[width=0.9\columnwidth]{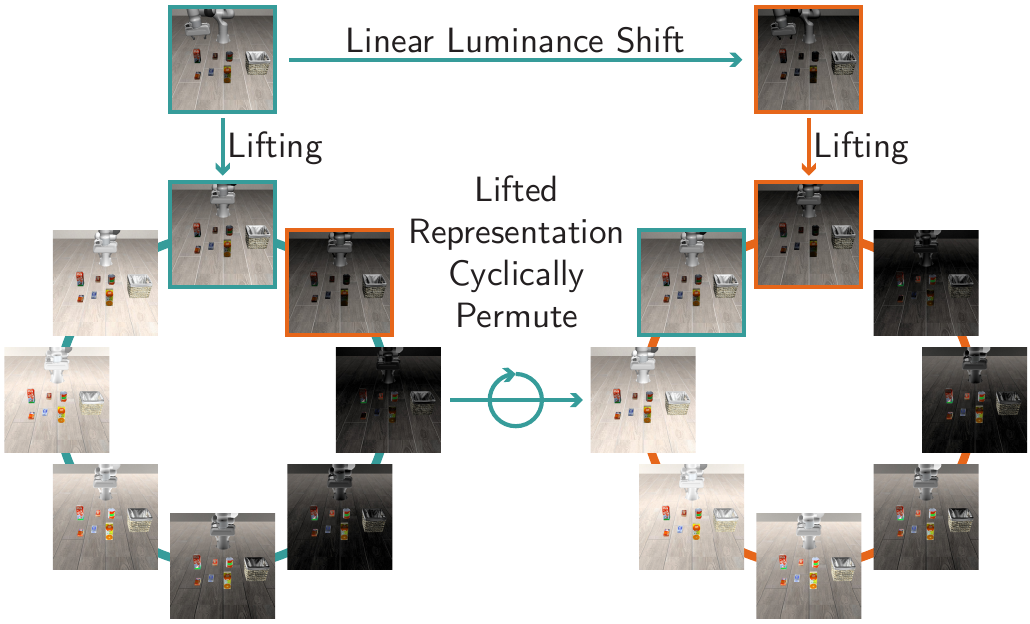}
    \caption{\textbf{CARROT Luminance-Group Lifting.} A global luminance shift induces a predictable cyclic permutation of the lifted orbit. The corresponding lifted representations are highlighted.}
    \label{fig:T3DP_Cartoon}
\end{figure}

Conventional diffusion policies encode images with planar convolutional networks.
These layers share filters across image locations, but they impose no corresponding structure on how the representation should change when image luminance changes.
CARROT replaces the standard visual encoder with a photometric group convolutional network based on the hypertoroidal construction of~\citet{yang2026hypertoroidal}.
While the general construction admits independent actions on all three HSL channels, we formulate the encoder using the luminance subgroup $L_{R}$.
\paragraph{Luminance Group and Group Action.}
Luminance $\tilde{L}$ is typically defined on the interval $I=\left[0,1\right]$, We lift it to the double-covered luminance $L$ with circular coordinates $\theta_{l}\in\left[0, 2\pi\right]$. 
The resulting covering map
\begin{equation}
    \mathcal{C}\left(\theta\right) = \left\lvert\left(\left(\nicefrac{\theta}{\pi}-1\right)\mathrm{mod}\,2\right)-1\right\rvert,
\end{equation}
traverses the luminance interval once in each direction around the circle.
We call the half of the circle with $\theta\in\left[0,\pi\right]$ the lower branch, on which $\mathcal{C}\left(\theta\right)=\theta/\pi$ rises from 0 to 1, and the half with $\theta\in\left[\pi,2\pi\right]$ the upper branch, on which it returns from 1 to 0.
Each luminance value therefore has two representatives on the circle, allowing luminance transformations to act as invertible circular shifts before projection back onto $I=\left[0,1\right]$.
A uniform discretization of order $R$ of this circle gives the luminance group with underlying set
\begin{equation}
    L_{R}=\left\{l_{i}=\nicefrac{2\pi i}{R}\mid i=0,\ldots,R-1\right\},
\end{equation}
and binary operator $l_{i}\cdot l_{j} = \left(l_{i}+l_{j}\right)\bmod 2\pi$.
The group acts on the circle by the same rotation $\theta_l\mapsto\left(\theta_{l}+l_{i}\right)\bmod 2\pi$, which satisfies the identity and compatibility axioms by construction.
For an input image in HSL color space $x=\left(x_{h},x_{s},x_{l}\right)$, let $\theta_{l}$ denote its corresponding coordinate on the double covered luminance space, taken on the lower branch as $\theta_{l}=\pi x_{l}\in\left[0,\pi\right]$, on which $\mathcal{C}$ is invertible.
The action of $l_{i}\in L_{R}$ on $x$ is defined as
\begin{equation}
    \varphi_{l}\left(l_{i},x\right) = \left(x_{h},\; x_{s},\; \mathcal{C}\left(\left(\theta_{l}+l_{i}\right)\bmod 2\pi\right)\right).
\end{equation}
The shift is invertible on the circular space $L$, while the covering map returns the transformed luminance to its valid bounded interval without endpoint clipping.
On images, this action agrees with the rotation on $L$ whenever every pixel remains on the lower branch, which holds for the uniform shifts $x_{l}\mapsto x_{l}+\delta_{i}$ that keep every pixel in $I$, with $\delta_{i}\in\left(-1,1\right]$ congruent to $2i/R$ modulo $2$. 
We call these modeled luminance shifts.
On them, the action on $L$ restricts to a partial group action on images~\citep{exel1998partial}, under which the layers that follow are exactly equivariant.
Translating a finite image without wrapping behaves the same way, since a shift is invertible only when no content leaves the frame.
For a function $f=\left(f_{1},...,f_{R}\right)$ defined on the discretized luminance group, the corresponding action of $l_{i}\in L_{R}$ is a cyclic permutation
\begin{equation}
    \phi_{l}\left(l_{i},f\right) = \left(f_{\left(1+i\right)\mathrm{mod}\,R},\ldots,f_{\left(R+i\right)\mathrm{mod}\,R}\right).
\end{equation}
\paragraph{Lifting and Group Convolutions.}
Group convolution is defined between functions on a group, so the observation must first be mapped onto discretized luminance group $\mathrm{L}_{R}$. 
The lifting layer evaluates the group action on the input at every group element,
\begin{equation}
    f^{0}\left(l_{i}\right)=\varphi_{l}\left(l_{i},x\right), \qquad l_{i}\in \mathrm{L}_{R},
\end{equation}
producing a lifted representation indexed by luminance transformations.
Since every element is computed from the same lower-branch coordinate, a modeled luminance shift by $l_{j}$ cyclically permutes the lifted representation, replacing each $f^{0}\left(l_{i}\right)$ with $f^{0}\left(l_{i}\cdot l_{j}\right)$.
We show an example of the lifted input-image orbit in Figure~\ref{fig:T3DP_Cartoon}.
Subsequent layers convolve this representation with filters defined on the group,
\begin{equation}
    \left[f^{\ell}\ast\psi_{i}^{\ell}\right]\left(a\right) = \sum_{r\in L_{R}}\sum_{k=1}^{K_{\ell}} f_{k}^{\ell}\left(r\right)\psi_{i,k}^{\ell}\left(r^{-1}a\right),
\end{equation}
which is equivariant to the luminance action.
To match the parameter count of the baseline Diffusion Policy encoder, we reduce each group-convolution layer’s channel width by a factor of $1/\sqrt{R}$.
An observation transformed by a modeled luminance shift therefore produces the same feature values up to a cyclic permutation of the luminance axis. 
Preserving this structure throughout the encoder requires every intermediate operation to commute with the same group action. 
Pointwise nonlinearities, residual sums, and spatial pooling and strides act identically at every luminance index and therefore commute with it.
We next define a normalization layer that shares its statistics across the complete luminance orbit and therefore maintains equivariance.
\subsection{Group Equivariant Group Norm}\label{subsec:method_instance_norm}
At each equivariant layer $\ell$, the feature map $f^{\ell}$ contains $K_{\ell}$ channels, each indexed by the luminance group $L_{R}$ and the spatial grid $\Omega_{\ell}$.
Diffusion Policy~\citep{chi2025diffusion} normalizes its ResNet18 encoder with group normalization~\citep{wu2018group} in place of batch normalization, and we extend this normalization to the luminance orbit.
The channels are partitioned into $G_{\ell}$ groups $\mathcal{K}_{1},\ldots,\mathcal{K}_{G_{\ell}}$ of equal size, each spanning $N_{\ell}=\lvert\mathcal{K}_{m}\rvert R\lvert\Omega_{\ell}\rvert$ entries of the feature map.
For each input, the statistics of group $\mathcal{K}_{m}$ are computed jointly over its channels, all luminance group elements, and all spatial locations,
\begin{gather}
    \mu_{m}^{\ell} = \frac{1}{N_{\ell}}\sum_{k\in\mathcal{K}_{m}}\sum_{r\in L_{R}}\sum_{p\in\Omega_{\ell}}f^{\ell}_{k}\left(r, p\right),\\
    \left(\sigma_{m}^{\ell}\right)^{2} = \frac{1}{N_{\ell}}\sum_{k\in\mathcal{K}_{m}}\sum_{r\in L_{R}}\sum_{p\in\Omega_{\ell}}\left(f^{\ell}_{k}\left(r, p\right) - \mu_{m}^{\ell}\right)^{2}.
\end{gather}
We define the group equivariant group normalization as
\begin{equation}
    \mathrm{GGN}\left(f^{\ell}\right)_{k}\left(r, p\right) = \gamma_{k}^{\ell}\frac{f_{k}^{\ell}\left(r, p\right)-\mu_{m}^{\ell}}{\sqrt{\left(\sigma_{m}^{\ell}\right)^{2}+\epsilon}}+\beta_{k}^{\ell},
\end{equation}
with $k\in\mathcal{K}_{m}$ and where $\gamma_{k}^{\ell}$ and $\beta_{k}^{\ell}$ are learned per channel and shared across the lifting orbit $r\in L_{R}$.
The action $\phi_{l}\left(l_{j},\cdot\right)$ permutes the luminance index of every channel.
Summation over the complete orbit therefore leaves the mean and variance of every group unchanged, and since $\gamma_{k}^{\ell}$ and $\beta_{k}^{\ell}$ do not depend on $r$, the layer preserves equivariance
\begin{equation}
    \mathrm{GGN}\left(\phi_{l}\left(l_{j}, f^{\ell}\right)\right) = \phi_{l}\left(l_{j},\mathrm{GGN}\left(f^{\ell}\right)\right).
\end{equation}
Group equivariant normalization therefore preserves the cyclic structure established by the group convolutions throughout the encoder.
At the encoder output, we pool this remaining luminance coordinate while retaining the spatial information needed to condition the diffusion policy.
\subsection{Invariant Spatial Keypoint Readout}\label{subsec:method_softmax}
Earlier color equivariant networks were developed primarily for classification, where the final readout can discard spatial structure~\citep{lengyel2023color,yang2025lcer,yang2026hypertoroidal}, whereas manipulation depends on where task relevant visual evidence appears in the image.
We therefore combine pooling over the luminance group with the spatial softmax~\citep{finn2016deep} used by diffusion policy encoders.

At the final equivariant layer $\ell_{\mathrm{out}}$, we maxpool each feature channel over its complete luminance orbit independently at every spatial location of $\Omega_{\ell_{\mathrm{out}}}$ such that
\begin{equation}
    \bar{f}_{i}\left(p\right) = \max_{r\in L_{R}}f_{i}^{\ell_{\mathrm{out}}}\left(x\right)\left(r, p\right).
\end{equation}
A modeled luminance shift only reorders the values along the luminance axis, so the pooled feature map satisfies
\begin{equation}
    \bar{f}_{i}\left(\varphi_{l}\left(l_{j}, x\right)\right)\left(p\right) = \bar{f}_{i}\left(x\right)\left(p\right),
\end{equation}
and retains the spatial organization of the scene.
Spatial softmax projects the invariant feature map into $K_{\mathrm{kp}}$ keypoints
\begin{equation}
    \kappa_{i} = \sum_{p\in\Omega_{\ell_{\mathrm{out}}}}\frac{\exp\left(\bar{f}_{i}\left(p\right)/\tau\right)}{\sum_{q\in\Omega_{\ell_{\mathrm{out}}}}\exp\left(\bar{f}_{i}\left(q\right)/\tau\right)}\nu\left(p\right),
\end{equation}
where $\nu\left(p\right)\in\left[-1, 1\right]^{2}$ is the normalized coordinate of spatial cell $p$ and $\tau>0$ is the temperature.
The visual encoder concatenates the resulting keypoints
\begin{equation}
    E_{\psi}\left(x\right) = \left[\kappa_{1},\ldots,\kappa_{K_{\mathrm{kp}}}\right].
\end{equation}
Because the keypoint maps are derived from an invariant feature map, the complete visual representation satisfies, for every modeled luminance shift,
\begin{equation}
    E_{\psi}\left(\varphi_{l}\left(l_{j}, x\right)\right) = E_{\psi}\left(x\right).
\end{equation}
The readout therefore supplies the denoiser with the same conditioning vector $c_{t}$ under any modeled luminance shift, and hence the same distribution, without discarding the spatial info needed for control.
For perturbations outside this set, the change in conditioning is bounded in proportion to how far the lifted observation lies from the lifted orbit, a quantity we define in Appendix~\ref{apx:quant_invariance} and measure in Appendix~\ref{apx:orbit_distance}.
\section{Experiments}
We evaluate whether the equivariance of the preceding section yields robustness as photometric shifts progressively depart from the modeled action.
Unperturbed LIBERO~\citep{liu2023libero} follows the training distribution, while luminance shifted LIBERO applies a controlled global shift with clipping.
LIBERO Plus~\citep{fei2025libero} adds spatially varying illumination intensity, direction, hue, and shadow.
On a Franka FR3 robot, physical changes in lighting intensity and direction also pass through sensor noise and a nonlinear camera response.
We compare CARROT with Diffusion Policy and methods that address photometric shift through augmentation~\citep{hendrycks2021many, zini2022planckian}, observation preprocessing~\citep{hu2025stem}, or restoration~\citep{wei2018deep, guo2020zero}, with implementation details in Appendix~\ref{apx:baselines}.
The real world evaluation further includes point cloud policies~\citep{ze20243d}.
Unless stated otherwise, single task policies are trained on demonstrations collected under nominal lighting, 50 per task in simulation and 40 in real world experiments, and evaluated without adaptation.
Simulated results report the mean and standard deviation over three training seeds.
\subsection{LIBERO}
\begin{table}[b]
    \centering
    \caption{\textbf{Unperturbed LIBERO Success Rate.} Under nominal lighting, CARROT matches Diffusion Policy and outperforms all preprocessing and restoration baselines.}
    \label{tab:libero_vanilla}
    \small
    \setlength{\tabcolsep}{2pt}
    \renewcommand{\arraystretch}{0.9}
    \begin{tabular}{lcccc}
        \toprule
        Network & Goal & Obj. & Spa. & Avg.\\
        \midrule
        Diffusion Policy & $90.5{\scriptstyle\pm1.8}$ & $\underline{\smash{86.5{\scriptstyle\pm4.0}}}$ & $84.7{\scriptstyle\pm2.3}$ & $\underline{\smash{87.3{\scriptstyle\pm1.2}}}$\\
        RetinexNet & $63.0{\scriptstyle\pm8.0}$ & $55.3{\scriptstyle\pm6.5}$ & $54.2{\scriptstyle\pm5.2}$ & $57.5{\scriptstyle\pm1.5}$\\
        Zero-DCE & $64.0{\scriptstyle\pm5.7}$ & $55.3{\scriptstyle\pm9.1}$ & $55.8{\scriptstyle\pm5.3}$ & $58.4{\scriptstyle\pm6.4}$\\
        Stem-OB & $\underline{\smash{91.8{\scriptstyle\pm1.0}}}$ & $83.0{\scriptstyle\pm3.6}$ & $\underline{\smash{85.0{\scriptstyle\pm0.5}}}$ & $86.6{\scriptstyle\pm1.3}$\\
        \midrule
        CARROT & $\mathbf{92.0{\scriptstyle\pm0.5}}$ & $\mathbf{87.0{\scriptstyle\pm2.3}}$ & $\mathbf{85.2{\scriptstyle\pm1.2}}$ & $\mathbf{88.1{\scriptstyle\pm1.1}}$\\
        \bottomrule
    \end{tabular}
\end{table}
\begin{figure}[t]
    \centering
    \includegraphics[width=\columnwidth]{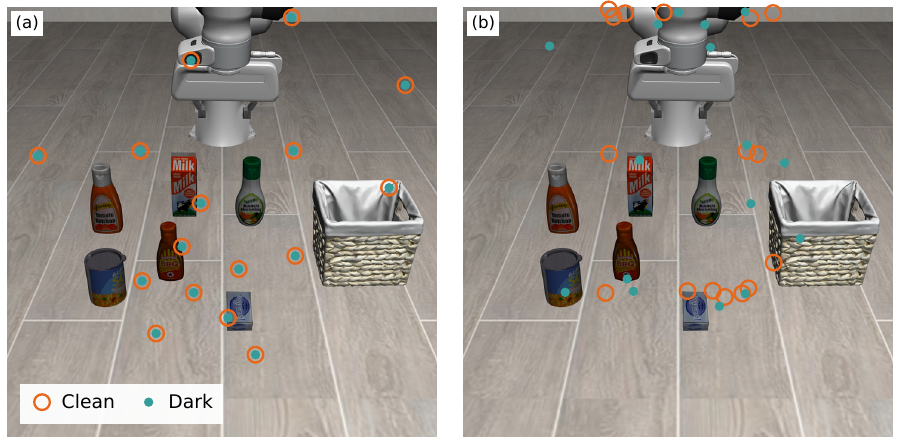}
    \caption{\textbf{Spatial Keypoint Stability.} We overlay the keypoints of \textbf{(a)} CARROT and \textbf{(b)} Diffusion Policy on a clean frame, extracted from the clean observation (hollow) and from the same frame darkened by a modeled luminance shift (filled). CARROT keypoints stays inside its clean counterpart, while Diffusion Policy's drift.}
    \label{fig:T3DP_Keypoint}
\end{figure}
\begin{table}[b]
    \centering
    \caption{\textbf{Luminance-Shifted LIBERO Success Rate.} Under a uniform luminance shift of $-0.25$ with clipping, CARROT outperforms Diffusion Policy, augmentation schemes, preprocessing, and restoration baselines.}
    \label{tab:libero_luminance}
    \small
    \setlength{\tabcolsep}{2pt}
    \renewcommand{\arraystretch}{0.9}
    \begin{tabular}{lcccc}
        \toprule
        Network & Goal & Obj. & Spa. & Avg.\\
        \midrule
        Diffusion Policy & $50.3{\scriptstyle\pm4.0}$ & $50.8{\scriptstyle\pm3.9}$ & $31.8{\scriptstyle\pm5.5}$ & $44.3{\scriptstyle\pm2.3}$\\
        DP-Jitter & $66.7{\scriptstyle\pm4.3}$ & $47.8{\scriptstyle\pm6.3}$ & $50.7{\scriptstyle\pm5.9}$ & $55.1{\scriptstyle\pm3.0}$\\
        DP-Planck & $56.5{\scriptstyle\pm1.3}$ & $49.5{\scriptstyle\pm9.1}$ & $36.0{\scriptstyle\pm6.9}$ & $47.3{\scriptstyle\pm2.3}$\\
        DP-DeepAug & $75.3{\scriptstyle\pm5.4}$ & $37.7{\scriptstyle\pm7.8}$ & $52.8{\scriptstyle\pm3.5}$ & $55.3{\scriptstyle\pm3.3}$\\
        RetinexNet & $72.7{\scriptstyle\pm4.5}$ & $67.7{\scriptstyle\pm4.5}$ & $\underline{\smash{80.8{\scriptstyle\pm1.0}}}$ & $73.7{\scriptstyle\pm2.2}$\\
        Zero-DCE & $72.8{\scriptstyle\pm0.8}$ & $\underline{\smash{73.5{\scriptstyle\pm1.5}}}$ & $76.3{\scriptstyle\pm1.3}$ & $\underline{\smash{74.2{\scriptstyle\pm0.4}}}$\\
        Stem-OB & $\underline{\smash{73.7{\scriptstyle\pm4.5}}}$ & $46.7{\scriptstyle\pm9.0}$ & $39.2{\scriptstyle\pm6.8}$ & $53.2{\scriptstyle\pm6.0}$\\
        \midrule
        CARROT & $\mathbf{77.8{\scriptstyle\pm0.8}}$ & $\mathbf{74.7{\scriptstyle\pm3.5}}$ & $\mathbf{83.8{\scriptstyle\pm4.6}}$ & $\mathbf{78.8{\scriptstyle\pm0.9}}$\\
        \bottomrule
    \end{tabular}
\end{table}
\paragraph{Unperturbed Performance.}
We first test whether the luminance equivariant encoder introduces a performance tradeoff on clean observations.
In Table~\ref{tab:libero_vanilla}, CARROT matches vanilla Diffusion Policy in overall success rate, indicating that the architectural constraint carries no cost under the training distribution.
In contrast, the restoration front ends substantially reduce clean performance, showing that modifying an already well exposed input can itself disrupt the visual conditioning expected by the policy.
\paragraph{Modeled Luminance Shift.}
Under a modeled luminance shift, the feature maps of CARROT cyclically permute along the luminance axis, and the invariant keypoint readout removes this permutation, so the denoiser receives identical conditioning.
In Figure~\ref{fig:T3DP_Keypoint}, CARROT keypoints accordingly coincide before and after the shift, while Diffusion Policy keypoints drift.
Appendix~\ref{apx:orbit_distance} visualizes the permutation in Figure~\ref{fig:feature_map_permutation} and confirms the invariance numerically in Table~\ref{tab:exact_invariance}.
\paragraph{Clipped Luminance Shift.}
We next introduce a controlled luminance transformation that closely follows the symmetry of the equivariant encoder.
For an input pixel $x=\left(x_{h},x_{s},x_{l}\right)$, the luminance channel is reduced by 0.25 and clipped to the valid image range
\begin{equation}
    x'= \left(x_{h},x_{s},\mathrm{max}\left(x_{l}-0.25, 0.0\right)\right).\label{eqn:libero_saturation_shift}
\end{equation}
A reduction of 0.25 coincides with an element of the discretized luminance group used by CARROT.
Clipping nevertheless breaks the exact group action by mapping luminance values below 0.25 to zero.
We therefore treat this evaluation as a near orbit shift rather than an exact test of the invariance guarantee.
Table~\ref{tab:orbit_distance_benchmark} in Appendix~\ref{apx:orbit_distance} shows that the resulting residual orbit distance lies entirely on clipped pixels and that CARROT keypoints drift less than those of Diffusion Policy.

Following the perturbation protocol of LIBERO Plus~\citep{fei2025libero}, we perturb only the observations.
Scene geometry, task state, and low dimensional robot observations remain fixed, and the same transformation is applied to both cameras.

In Table~\ref{tab:libero_luminance}, CARROT achieves the highest success in every suite under the shifted observations.
Relative to the unperturbed condition, CARROT retains approximately 89\% of its success, while vanilla Diffusion Policy retains approximately 51\%.
Augmentation and observation preprocessing improve individual suites but not consistently across the benchmark.
Restoration remains competitive under darkening, but Table~\ref{tab:libero_vanilla} shows that this robustness comes with a substantial clean performance cost.
\subsection{LIBERO-Plus Lighting}
\begin{table}[b]
    \centering
    \caption{\textbf{LIBERO-Plus Lighting Success Rate.} Under LIBERO-Plus lighting perturbations, CARROT outperforms Diffusion Policy, augmentation schemes, preprocessing, and restoration baselines.}
    \label{tab:libero_plus_lighting}
    \small
    \setlength{\tabcolsep}{2pt}
    \renewcommand{\arraystretch}{0.9}
    \begin{tabular}{lcccc}
        \toprule
        Network & Goal & Obj. & Spa. & Avg.\\
        \midrule
        Diffusion Policy & $48.7{\scriptstyle\pm1.7}$ & $46.1{\scriptstyle\pm1.4}$ & $28.0{\scriptstyle\pm9.0}$ & $40.9{\scriptstyle\pm3.2}$\\
        DP-Jitter & $54.2{\scriptstyle\pm8.3}$ & $49.1{\scriptstyle\pm10.4}$ & $50.8{\scriptstyle\pm2.9}$ & $51.3{\scriptstyle\pm4.6}$\\
        DP-Planck & $56.1{\scriptstyle\pm1.5}$ & $39.2{\scriptstyle\pm4.3}$ & $40.5{\scriptstyle\pm3.3}$ & $45.2{\scriptstyle\pm3.0}$\\
        DP-DeepAug & $58.1{\scriptstyle\pm6.4}$ & $36.6{\scriptstyle\pm6.3}$ & $52.2{\scriptstyle\pm4.3}$ & $48.9{\scriptstyle\pm4.3}$\\
        RetinexNet & $38.8{\scriptstyle\pm5.9}$ & $43.1{\scriptstyle\pm6.0}$ & $51.6{\scriptstyle\pm6.7}$ & $44.5{\scriptstyle\pm2.6}$\\
        Zero-DCE & $48.5{\scriptstyle\pm4.2}$ & $\underline{\smash{60.4{\scriptstyle\pm7.7}}}$ & $\underline{\smash{61.9{\scriptstyle\pm10.0}}}$ & $\underline{\smash{56.9{\scriptstyle\pm6.4}}}$\\
        Stem-OB & $\underline{\smash{64.7{\scriptstyle\pm3.8}}}$ & $33.9{\scriptstyle\pm3.3}$ & $43.5{\scriptstyle\pm10.7}$ & $47.4{\scriptstyle\pm3.8}$\\
        \midrule
        CARROT & $\mathbf{71.4{\scriptstyle\pm8.8}}$ & $\mathbf{67.1{\scriptstyle\pm6.3}}$ & $\mathbf{71.7{\scriptstyle\pm6.5}}$ & $\mathbf{70.2{\scriptstyle\pm4.0}}$\\
        \bottomrule
    \end{tabular}
\end{table}
\paragraph{Spatially Varying Lighting Shift.}
We evaluate each single task policy under the LIBERO Plus~\citep{fei2025libero} lighting variants of its task, so rendered illumination is the only source of shift.
Figure~\ref{fig:T3DP_LIBEROPlusDatasetDemo} in the appendix illustrates these variants, and Table~\ref{tab:orbit_distance_benchmark} in Appendix~\ref{apx:orbit_distance} places them considerably further from the luminance orbit than luminance shifted LIBERO.
In Table~\ref{tab:libero_plus_lighting}, CARROT achieves the highest success in every suite and retains approximately 80\% of its unperturbed success, compared with approximately 47\% for Diffusion Policy.
Augmentation, restoration, and observation preprocessing improve individual suites but do not match the consistent gains of CARROT across the benchmark.
Restoration in particular loses much of the advantage it held under uniform darkening.
The luminance equivariant representation therefore remains useful beyond the modeled transformation.
Although no encoder is invariant to these perturbations, Table~\ref{tab:libero_plus_keypoint_error} in Appendix~\ref{appendix:exp_libero_plus} shows that CARROT keypoints drift the least, linking the success gains to a smaller change in the visual conditioning supplied to the denoiser.
%
\begin{table}[t]
    \centering
    \caption{\textbf{Language-Conditioned LIBERO-Plus Lighting Success Rate.} We report suite-level success on the Goal, Object, and Spatial suites of LIBERO evaluated under lighting perturbations defined in LIBERO-Plus. CARROT outperforms vanilla Diffusion Policy and restoration baselines under language-conditioned manipulation.}
    \label{tab:libero_plus_language}
    \small
    \setlength{\tabcolsep}{2pt}
    \renewcommand{\arraystretch}{0.9}
    \begin{tabular}{lcccc}
        \toprule
        Network & Goal & Obj. & Spa. & Avg.\\
        \midrule
        Diffusion Policy & $30.5{\scriptstyle\pm5.4}$ & $7.4{\scriptstyle\pm6.9}$ & $20.1{\scriptstyle\pm4.9}$ & $19.3{\scriptstyle\pm5.2}$\\
        RetinexNet & $25.7{\scriptstyle\pm4.3}$ & $45.2{\scriptstyle\pm12.1}$ & $26.2{\scriptstyle\pm9.9}$ & $32.4{\scriptstyle\pm2.9}$\\
        Zero-DCE & $\underline{\smash{50.2{\scriptstyle\pm10.2}}}$ & $\underline{\smash{61.9{\scriptstyle\pm2.0}}}$ & $\underline{\smash{40.8{\scriptstyle\pm7.7}}}$ & $\underline{\smash{51.0{\scriptstyle\pm4.2}}}$\\
        Stem-OB & $34.9{\scriptstyle\pm2.8}$ & $27.6{\scriptstyle\pm6.7}$ & $31.1{\scriptstyle\pm6.3}$ & $31.2{\scriptstyle\pm0.8}$\\
        \midrule
        CARROT & $\mathbf{70.9{\scriptstyle\pm6.5}}$ & $\mathbf{64.5{\scriptstyle\pm4.5}}$ & $\mathbf{56.5{\scriptstyle\pm1.6}}$ & $\mathbf{63.9{\scriptstyle\pm3.9}}$\\
        \bottomrule
    \end{tabular}
\end{table}
\begin{table}[b]
    \centering
    \caption{\textbf{LIBERO-Plus Invariance Mechanisms.} We report the success rate, and training time and memory usage relative to CARROT on the Spatial suite under LIBERO-Plus lighting perturbations. CARROT outperforms shared-convolutions, on-orbit augmentation, and test-time ensembling in success rate, while striking a balance in terms of training time and memory. Implementation details are provided in Appendix~\ref{apx:ablation}.}
    \label{tab:libero_plus_ablation}
    \small
    \setlength{\tabcolsep}{4pt}
    \renewcommand{\arraystretch}{0.9}
    \begin{tabular}{lcccc}
        \toprule
        Variant & Spa. & Time & Mem.\\
        \midrule
        CARROT & $\mathbf{71.7{\scriptstyle\pm6.5}}$ & $\times1.00$ & $\times1.00$\\
        \midrule
        Shared-Conv & $62.3{\scriptstyle\pm5.4}$ & $\times1.49$ & $\times1.66$\\
        Augmentation & $57.1{\scriptstyle\pm3.2}$ & $\times3.22$ & $\times0.66$\\
        Ensembling & $42.1{\scriptstyle\pm5.1}$ & $\times0.44$ & $\times0.66$\\
        \bottomrule
    \end{tabular}
\end{table}
\paragraph{Lighting Shift with Language Conditioning.}
Single task policies fix task identity by construction, so we next test whether luminance equivariance remains beneficial when one policy must distinguish tasks through language.
For each suite, we train one policy on all ten tasks, concatenating the visual keypoints with the BERT~\citep{devlin2019bert} task embedding provided by LIBERO in the conditioning supplied to the denoiser.
The resulting policies are evaluated under the same LIBERO Plus lighting variants.
Language identifies the task but does not remove the photometric shift from the visual observation.
In Table~\ref{tab:libero_plus_language}, CARROT achieves the highest success in every suite and more than triples the overall success of Diffusion Policy.
Equivariance therefore remains beneficial when visual and language conditioning are learned jointly.
\paragraph{Alternative Invariance Mechanisms.}
Under the same LIBERO-Plus lighting perturbations, Table~\ref{tab:libero_plus_ablation} compares CARROT on the Spatial suite with three alternative invariance mechanisms, with implementation details in Appendix~\ref{apx:ablation}. 
A shared-convolution encoder applies common planar filters to every element of the lifted orbit before pooling; on-orbit augmentation trains a planar encoder on observations transformed by the group; and test-time ensembling pools Diffusion Policy over the orbit at evaluation.
CARROT achieves the highest success rate and requires less training time than shared convolutions and augmentation, the two alternatives that come closest.
Ensembling adds no additional training requirement but performs worst.
Access to the luminance orbit alone therefore does not account for the gains of CARROT, which arise from building the invariance into the encoder through equivariant group convolutions.
\paragraph{Group Order.}
\begin{figure}[t]
    \centering
    \includegraphics[width=\columnwidth]{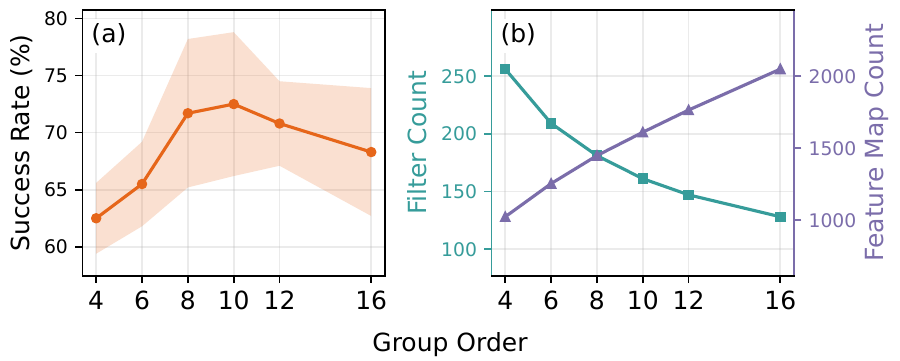}
    \caption{\textbf{Group Order.} We report \textbf{(a)} Spatial suite success under LIBERO-Plus lighting, with one standard deviation across three seeds shaded, and \textbf{(b)} the filter and feature map counts of the final encoder stage, both against the group order $R$ at matched parameter count. Success rises up to $R=8$ and remains within seed variation for larger orders, which leave fewer filters for each group element.}
    \label{fig:CARROT_order}
\end{figure}
CARROT uses $R=8$ in all reported experiments.
Figure~\ref{fig:CARROT_order} retrains on the Spatial suite under LIBERO-Plus lighting for $R\in\left\{4,6,8,10,12,16\right\}$.
With the parameter count held fixed, a larger $R$ discretizes luminance more finely but leaves fewer filters for each group element, as shown in Figure~\ref{fig:CARROT_order}(b).
Success rises from $R=4$ to $R=8$, and the differences among larger orders remain within the variation across seeds, consistent with the reduced width per group element offsetting the finer discretization.
Every order remains well above Diffusion Policy in Table~\ref{tab:libero_plus_lighting}, so the gain of CARROT over Diffusion Policy does not depend on a particular choice of $R$.
\paragraph{Demonstration Scaling.}
\begin{figure}[b]
    \centering
    \includegraphics[width=\columnwidth]{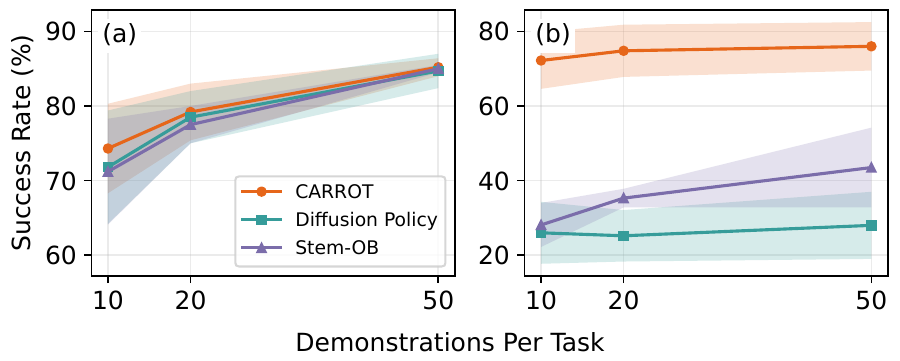}
    \caption{\textbf{Training Sample Efficiency.} We report Spatial suite success against the number of demonstrations per task on \textbf{(a)} unperturbed LIBERO and \textbf{(b)} LIBERO-Plus lighting, each under its own evaluation protocol. CARROT matches both baselines on clean observations at every data scale, while under lighting shift CARROT trained on 10 demonstrations exceeds both baselines trained on 50.}
    \label{fig:CARROT_sample}
\end{figure}
To examine how robustness scales with data, we train Spatial suite policies on 10, 20, and 50 demonstrations per task, with protocol details in Appendix~\ref{apx:demo_scaling}.
Figure 3 shows that CARROT matches Diffusion Policy and Stem-OB on unperturbed LIBERO at every data scale, so the equivariant encoder imposes no cost in sample efficiency.
Under LIBERO-Plus lighting, CARROT trained on 10 demonstrations already outperforms both baselines trained on 50.
Additional demonstrations collected under nominal lighting leave Diffusion Policy nearly unchanged and improve Stem-OB only gradually.
The illumination robustness of CARROT therefore comes from its encoder rather than from the amount of training data.
\subsection{Real World Manipulation Tasks}
The simulated evaluations alter illumination at the renderer, so every change in appearance originates from the modeled scene lighting.
Physical deployment adds sensor noise~\citep{healey1994radiometric,foi2008practical} and a nonlinear camera response~\cite{debevec2023recovering,grossberg2004modeling}, and a reduction in light reaches the policy through a transformation that the luminance group only approximates.
\begin{figure}[b]
    \centering
    \includegraphics[width=\linewidth]{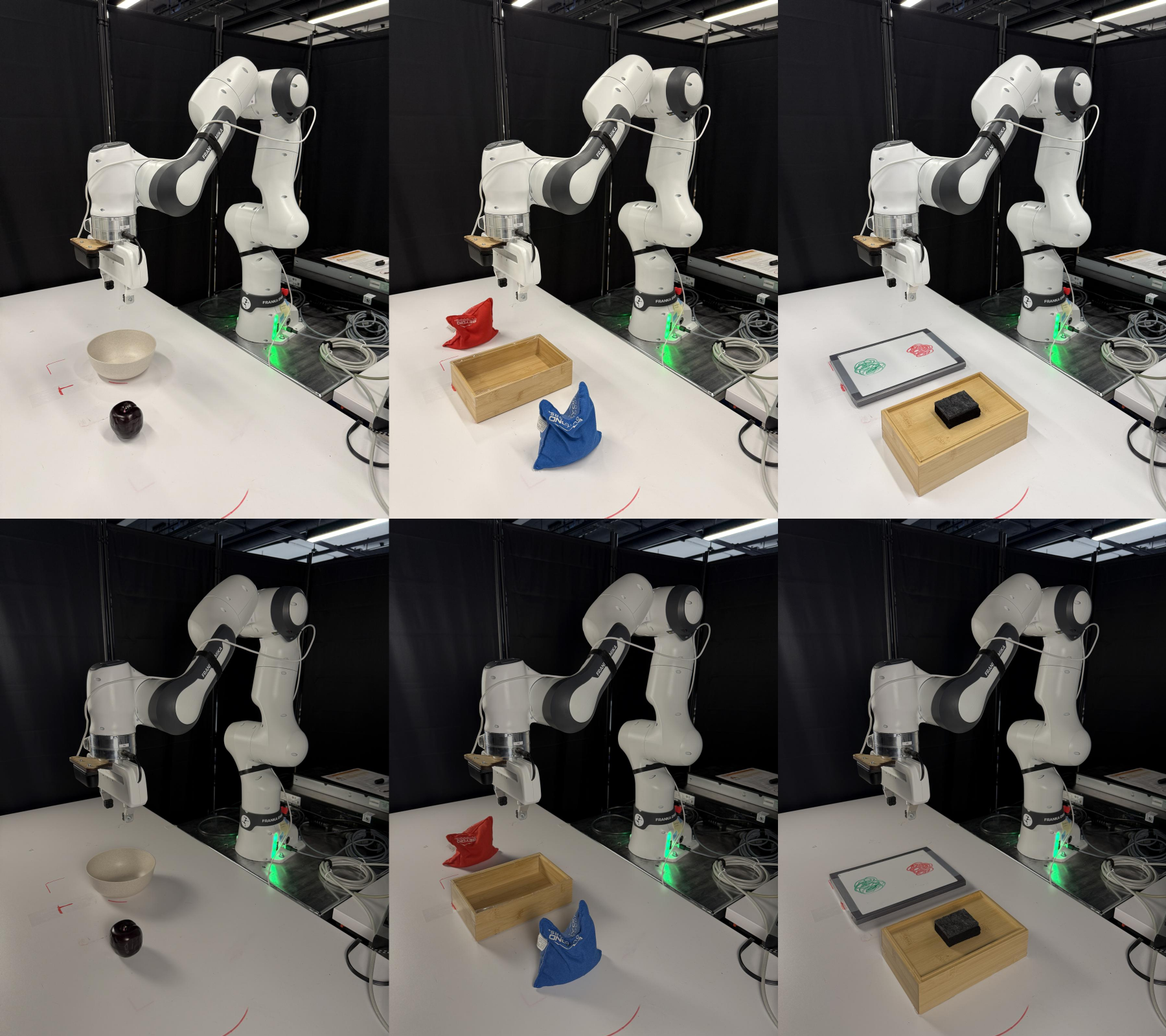}
    \caption{\textbf{Real world manipulation tasks.} Experimental setups for the (left) Cherry, (middle) Beanbag, and (right) Eraser manipulation tasks using a Franka FR3 robot. In the Beanbag and Eraser tasks, the robot must move the red beanbag into the tray and erase the red marking from the whiteboard. The top and bottom rows show the light and dark conditions, respectively.}
    \label{fig:CARROT_Demo}
\end{figure}
We evaluate CARROT on a Franka FR3 robot across three tasks under nominal and reduced lighting, shown in Figure~\ref{fig:CARROT_Demo}. These span pick and place (Fruit), color-based sorting (Bean), and contact-rich multi-step tool use (Eraser), of which only the latter two depend on color information to correctly identify the appropriate actions.
All policies are trained on 40 demonstrations per task collected under nominal lighting and evaluated without adaptation over 10 trials per condition.
Alongside vanilla Diffusion Policy, Zero-DCE, and Stem-OB, we compare against DP3~\citep{ze20243d}, whose point cloud input is insensitive to appearance but cannot resolve color, and its colored variant DP3-RGB.
\paragraph{Nominal Lighting.} 
\begin{table}[t]
    \centering
    \caption{\textbf{Real World Task Success Rate under Nominal Lighting.} CARROT matches Diffusion Policy on every task, indicating that the equivariance incurs no performance cost under clean lighting.}
    \label{tab:realworld_light}
    \small
    \renewcommand{\arraystretch}{0.9}
    \begin{tabular}{lcccc}
        \toprule
        Model & Fruit & Bean & Eraser & Avg.\\
        \midrule
        Diffusion Policy & $0.9$ & $0.9$ & $0.8$ & $\underline{\smash{0.87}}$\\
        Zero-DCE & $0.2$ & $0.1$ & $0.0$ & $0.10$\\
        Stem-OB & $0.8$ & $0.9$ & $0.8$ & $0.83$\\
        DP3 & $1.0$ & $0.3$ & $0.3$ & $0.53$\\
        DP3-RGB & $0.9$ & $0.9$ & $0.9$ & $\mathbf{0.90}$\\
        \midrule
        CARROT & $0.9$ & $0.9$ & $0.8$ & $\underline{\smash{0.87}}$\\
        \bottomrule
    \end{tabular}
\end{table}
Table~\ref{tab:realworld_light} reports success under the lighting in which the demonstrations were collected.
CARROT matches Diffusion Policy on every task, so the equivariant encoder carries no performance cost on real hardware.
Zero-DCE fails on most trials even without a lighting change, consistent with the simulated finding that enhancing an already well exposed observation disrupts the visual conditioning the policy expects.
DP3 succeeds on the Fruit task but falls behind on both color-sensitive tasks, while DP3-RGB closes this gap by adding color to depth, indicating that these tasks require chromatic information beyond scene geometry.
\paragraph{Reduced Lighting.} 
\begin{table}[b]
    \centering
    \caption{\textbf{Real World Task Success Rate under Reduced Lighting.} We report success rates under reduced lighting, along with the scene average luminance change $\Delta$-L relative to nominal lighting at the agent and wrist cameras. CARROT achieves the highest average success rate and leads on both color-sensitive tasks.}
    \label{tab:realworld_dark}
    \small
    \renewcommand{\arraystretch}{0.9}
    \begin{tabular}{lcccc}
        \toprule
        Model & Fruit & Bean & Eraser & Avg.\\
        \midrule
        Agent $\Delta$L & $-26\%$ & $-28\%$ & $-26\%$ & $-27\%$\\
        Wrist $\Delta$L & $-34\%$ & $-36\%$ & $-35\%$ & $-35\%$\\
        Color-Sensitive & \xmark & \cmark & \cmark & -\\
        \midrule
        Diffusion Policy & $0.2$ & $0.3$ & $0.2$ & $0.23$\\
        Zero-DCE & $0.3$ & $0.0$ & $0.3$ & $0.20$\\
        Stem-OB & $0.2$ & $0.1$ & $0.3$ & $0.20$\\
        DP3 & $\mathbf{0.9}$ & $\underline{\smash{0.4}}$ & $0.3$ & $\underline{\smash{0.53}}$\\
        DP3-RGB & $0.6$ & $0.3$ & $\underline{\smash{0.4}}$ & $0.43$\\
        \midrule
        CARROT & $\underline{\smash{0.7}}$ & $\mathbf{0.8}$ & $\mathbf{0.7}$ & $\mathbf{0.73}$\\
        \bottomrule
    \end{tabular}
\end{table}
The reduced lighting condition replaces the overhead laboratory lighting with dimmer studio lighting from the side, changing both the intensity and the direction of illumination while each task setup is held fixed.
Table~\ref{tab:realworld_dark} reports the resulting change in scene average luminance at both cameras alongside the success rates.
Diffusion Policy degrades on all three tasks, and neither Zero-DCE nor Stem-OB prevents this degradation.
DP3 retains its success on the Fruit task, since the depth observations underlying its point clouds change little under reduced visible light, but it remains limited on the color-sensitive tasks.
DP3-RGB loses most of its nominal advantage on these tasks once color is observed under reduced light.
CARROT achieves the highest average success and leads on both color-sensitive tasks, the setting in which neither geometry alone nor an unstructured RGB encoder suffices.
The gains extend from rendered illumination to real lighting changes on a physical robot, where the shift departs further from the modeled luminance action.
\paragraph{Inference Latency.} 
Table~\ref{tab:latency} reports parameter count and inference latency for the compared models.
CARROT modifies only the visual encoder, so its overhead relative to Diffusion Policy arises from propagating features over the luminance group, and at matched parameter count this adds a modest cost to the forward pass.
Because the equivariance is built into the encoder, CARROT requires no per-request preprocessing, and its total latency is below DP3, which processes a point cloud using FPS~\citep{qi2017pointnet++} for every observation.
\begin{table}[t]
    \centering
    \caption{\textbf{Parameter Count and Inference Latency.} Forward latency is at batch size one over 10 forward passes after warm-up. Total latency adds the per-request preprocessing before the forward pass, namely Zero-DCE enhancement and point-cloud construction. At matched parameter count, CARROT adds only a modest latency overhead over Diffusion Policy.}
    \small
    \renewcommand{\arraystretch}{0.9}
    \label{tab:latency}
    \begin{tabular}{lc*{2}{r@{}l}}
        \toprule
        \multirow{2}{*}[-0.5ex]{Model} & \multirow{2}{*}[-0.5ex]{Params} & \multicolumn{4}{c}{Latency}\\
        \cmidrule(lr){3-6}
        & & \multicolumn{2}{c}{Forward} & \multicolumn{2}{c}{Total}\\
        \midrule
        Diffusion Policy & $278.5$M & $56.2$&${\scriptstyle\pm0.03}$ & $56.2$&${\scriptstyle\pm0.03}$\\
        Zero-DCE & $278.5$M & $56.2$&${\scriptstyle\pm0.03}$ & $60.3$&${\scriptstyle\pm0.03}$\\
        Stem-OB & $278.5$M & $56.1$&${\scriptstyle\pm0.04}$ & $56.1$&${\scriptstyle\pm0.04}$\\
        DP3 & $255.2$M & $54.6$&${\scriptstyle\pm0.19}$ & $65.7$&${\scriptstyle\pm0.23}$\\
        DP3-RGB & $255.3$M & $54.4$&${\scriptstyle\pm0.04}$ & $65.6$&${\scriptstyle\pm0.14}$\\
        \midrule
        CARROT & $278.4$M & $61.1$&${\scriptstyle\pm0.08}$ & $61.1$&${\scriptstyle\pm0.08}$\\
        \bottomrule
    \end{tabular}
\end{table}
\section{Conclusion}
We introduce CARROT, a diffusion policy that employs the outputs from its luminance-equivariant vision backbone to obtain invariant spatial conditioning under all modeled shifts of global luminance, leaving the denoiser and the diffusion objective unchanged.
At matched parameter count, CARROT preserves unperturbed performance in simulation and on hardware.
Under luminance-shifted LIBERO, LIBERO-Plus lighting, and reduced lighting on real hardware, it achieves the highest average task success, including with language-conditioned policies.
Alternative invariance mechanisms over the same luminance orbit fall short of CARROT, so its gains arise from building luminance equivariance into the encoder rather than from access to the orbit alone.
Although the formal guarantee applies to the modeled global luminance action, the gains extend to spatially varying lighting, under which CARROT keypoints drift the least among the measured encoders, and to real lighting changes, supporting photometric equivariance as a practical alternative to augmentation, preprocessing, and restoration.
\paragraph{Limitations.}
Exact invariance holds only for the discrete global luminance shifts that keep every pixel in range, and changes in illumination color are not modeled. 
While the size of CARROT is matched to that of the baseline, lifting observations and propagating features over the group increases convolutional cost in proportion to the group size~\citep{cohen2016group}.
{\small \bibliographystyle{ieeenat_fullname}
\bibliography{ref}}

@article{mirjalili2025arro,
  title={Augmented reality for robots (arro): Pointing visuomotor policies towards visual robustness},
  author={Mirjalili, Reihaneh and J{\"u}lg, Tobias and Walter, Florian and Burgard, Wolfram},
  journal={IEEE Robotics and Automation Letters},
  year={2026},
  publisher={IEEE}
}

@article{wang2024equivariant,
  title={Equivariant diffusion policy},
  author={Wang, Dian and Hart, Stephen and Surovik, David and Kelestemur, Tarik and Huang, Haojie and Zhao, Haibo and Yeatman, Mark and Wang, Jiuguang and Walters, Robin and Platt, Robert},
  journal={arXiv preprint arXiv:2407.01812},
  year={2024}
}

@article{lengyel2023color,
  title={Color equivariant convolutional networks},
  author={Lengyel, Attila and Strafforello, Ombretta and Bruintjes, Robert-Jan and Gielisse, Alexander and van Gemert, Jan},
  journal={Advances in Neural Information Processing Systems},
  volume={36},
  pages={29831--29850},
  year={2023}
}

@inproceedings{yang2025lcer,
  title={Learning color equivariant representations},
  author={Yang, Yulong and O'Mahony, Felix and Allen-Blanchette, Christine},
  booktitle={International Conference on Learning Representations},
  volume={2025},
  pages={18164--18190},
  year={2025}
}

@article{yang2026hypertoroidal,
  title={A Hypertoroidal Covering for Perfect Color Equivariance},
  author={Yang, Yulong and Xu, Zhikun and Li, Yaojun and Allen-Blanchette, Christine},
  journal={arXiv preprint arXiv:2603.04256},
  year={2026}
}

@article{chi2025diffusion,
  title={Diffusion policy: Visuomotor policy learning via action diffusion},
  author={Chi, Cheng and Xu, Zhenjia and Feng, Siyuan and Cousineau, Eric and Du, Yilun and Burchfiel, Benjamin and Tedrake, Russ and Song, Shuran},
  journal={The International Journal of Robotics Research},
  volume={44},
  number={10-11},
  pages={1684--1704},
  year={2025},
  publisher={Sage Publications Sage UK: London, England}
}

@article{ze20243d,
  title={3d diffusion policy: Generalizable visuomotor policy learning via simple 3d representations},
  author={Ze, Yanjie and Zhang, Gu and Zhang, Kangning and Hu, Chenyuan and Wang, Muhan and Xu, Huazhe},
  journal={arXiv preprint arXiv:2403.03954},
  year={2024}
}

@article{huang2025spatial,
  title={Spatial robograsp: Generalized robotic grasping control policy},
  author={Huang, Yiqi and Davies, Travis and Yan, Jiahuan and Sun, Jiankai and Chen, Xiang and Hu, Luhui},
  journal={arXiv preprint arXiv:2505.20814},
  year={2025}
}

@inproceedings{hu2025stem,
  title={Stem-OB: Generalizable visual imitation learning with stem-like convergent observation through diffusion inversion},
  author={Hu, Kaizhe and Rui, Zihang and He, Yao and Liu, Yuyao and Hua, Pu and Xu, Huazhe},
  booktitle={International Conference on Learning Representations},
  volume={2025},
  pages={7340--7367},
  year={2025}
}

@inproceedings{wang2024gendp,
  title={GenDP: 3D Semantic Fields for Category-Level Generalizable Diffusion Policy.},
  author={Wang, Yixuan and Yin, Guang and Huang, Binghao and Kelestemur, Tarik and Wang, Jiuguang and Li, Yunzhu},
  booktitle={CoRL},
  pages={4866--4878},
  year={2024}
}

@article{jia2026pointmappolicy,
  title={Pointmappolicy: Structured point cloud processing for multi-modal imitation learning},
  author={Jia, Xiaogang and Wang, Qian and Wang, Anrui and Wang, Han and Gyenes, Bal{\'a}zs and Gospodinov, Emiliyan and Jiang, Xinkai and Li, Ge and Zhou, Hongyi and Liao, Weiran and others},
  journal={Advances in Neural Information Processing Systems},
  volume={38},
  pages={160188--160216},
  year={2026}
}

@article{jin2026robolight,
  title={RoboLight: A Dataset with Linearly Composable Illumination for Robotic Manipulation},
  author={Jin, Shutong and Yang, Jin and Zahid, Muhammad and Pokorny, Florian T},
  journal={arXiv preprint arXiv:2603.04249},
  year={2026}
}

@inproceedings{wang2024rise,
  title={Rise: 3d perception makes real-world robot imitation simple and effective},
  author={Wang, Chenxi and Fang, Hongjie and Fang, Hao-Shu and Lu, Cewu},
  booktitle={2024 IEEE/RSJ International Conference on Intelligent Robots and Systems (IROS)},
  pages={2870--2877},
  year={2024},
  organization={IEEE}
}

@article{batra2025zero,
  title={Zero-shot visual generalization in robot manipulation},
  author={Batra, Sumeet and Sukhatme, Gaurav},
  journal={arXiv preprint arXiv:2505.11719},
  year={2025}
}

@inproceedings{ze2025generalizable,
  title={Generalizable humanoid manipulation with 3d diffusion policies},
  author={Ze, Yanjie and Chen, Zixuan and Wang, Wenhao and Chen, Tianyi and He, Xialin and Yuan, Ying and Peng, Xue Bin and Wu, Jiajun},
  booktitle={2025 IEEE/RSJ International Conference on Intelligent Robots and Systems (IROS)},
  pages={2873--2880},
  year={2025},
  organization={IEEE}
}

@article{ke20243d,
  title={3d diffuser actor: Policy diffusion with 3d scene representations},
  author={Ke, Tsung-Wei and Gkanatsios, Nikolaos and Fragkiadaki, Katerina},
  journal={arXiv preprint arXiv:2402.10885},
  year={2024}
}

@article{yang2024equibot,
  title={Equibot: Sim (3)-equivariant diffusion policy for generalizable and data efficient learning},
  author={Yang, Jingyun and Cao, Zi-ang and Deng, Congyue and Antonova, Rika and Song, Shuran and Bohg, Jeannette},
  journal={arXiv preprint arXiv:2407.01479},
  year={2024}
}

@article{bronstein2021geometric,
  title={Geometric deep learning: Grids, groups, graphs, geodesics, and gauges},
  author={Bronstein, Michael M and Bruna, Joan and Cohen, Taco and Veli{\v{c}}kovi{\'c}, Petar},
  journal={arXiv preprint arXiv:2104.13478},
  year={2021}
}

@inproceedings{cohen2016group,
  title={Group equivariant convolutional networks},
  author={Cohen, Taco and Welling, Max},
  booktitle={International conference on machine learning},
  pages={2990--2999},
  year={2016},
  organization={PMLR}
}

@inproceedings{worrall2017harmonic,
  title={Harmonic networks: Deep translation and rotation equivariance},
  author={Worrall, Daniel E and Garbin, Stephan J and Turmukhambetov, Daniyar and Brostow, Gabriel J},
  booktitle={Proceedings of the IEEE conference on computer vision and pattern recognition},
  pages={5028--5037},
  year={2017}
}

@inproceedings{kondor2018generalization,
  title={On the generalization of equivariance and convolution in neural networks to the action of compact groups},
  author={Kondor, Risi and Trivedi, Shubhendu},
  booktitle={International conference on machine learning},
  pages={2747--2755},
  year={2018},
  organization={PMLR}
}

@article{cohen2019general,
  title={A general theory of equivariant cnns on homogeneous spaces},
  author={Cohen, Taco S and Geiger, Mario and Weiler, Maurice},
  journal={Advances in neural information processing systems},
  volume={32},
  year={2019}
}

@inproceedings{esteves2018learning,
  title={Learning so (3) equivariant representations with spherical cnns},
  author={Esteves, Carlos and Allen-Blanchette, Christine and Makadia, Ameesh and Daniilidis, Kostas},
  booktitle={Proceedings of the european conference on computer vision (ECCV)},
  pages={52--68},
  year={2018}
}

@article{esteves2017polar,
  title={Polar transformer networks},
  author={Esteves, Carlos and Allen-Blanchette, Christine and Zhou, Xiaowei and Daniilidis, Kostas},
  journal={arXiv preprint arXiv:1709.01889},
  year={2017}
}

@article{worrall2019deep,
  title={Deep scale-spaces: Equivariance over scale},
  author={Worrall, Daniel and Welling, Max},
  journal={Advances in Neural Information Processing Systems},
  volume={32},
  year={2019}
}

@article{sun2023empowering,
  title={Empowering networks with scale and rotation equivariance using a similarity convolution},
  author={Sun, Zikai and Blu, Thierry},
  journal={arXiv preprint arXiv:2303.00326},
  year={2023}
}

@article{liu2023libero,
  title={Libero: Benchmarking knowledge transfer for lifelong robot learning},
  author={Liu, Bo and Zhu, Yifeng and Gao, Chongkai and Feng, Yihao and Liu, Qiang and Zhu, Yuke and Stone, Peter},
  journal={Advances in Neural Information Processing Systems},
  volume={36},
  pages={44776--44791},
  year={2023}
}

@article{thomas2018tensor,
  title={Tensor field networks: Rotation-and translation-equivariant neural networks for 3d point clouds},
  author={Thomas, Nathaniel and Smidt, Tess and Kearnes, Steven and Yang, Lusann and Li, Li and Kohlhoff, Kai and Riley, Patrick},
  journal={arXiv preprint arXiv:1802.08219},
  year={2018}
}

@inproceedings{esteves2019equivariant,
  title={Equivariant multi-view networks},
  author={Esteves, Carlos and Xu, Yinshuang and Allen-Blanchette, Christine and Daniilidis, Kostas},
  booktitle={Proceedings of the IEEE/CVF international conference on computer vision},
  pages={1568--1577},
  year={2019}
}

@inproceedings{cohen2019gauge,
  title={Gauge equivariant convolutional networks and the icosahedral CNN},
  author={Cohen, Taco and Weiler, Maurice and Kicanaoglu, Berkay and Welling, Max},
  booktitle={International conference on Machine learning},
  pages={1321--1330},
  year={2019},
  organization={PMLR}
}

@inproceedings{buhrmester2019evaluating,
  title={Evaluating the impact of color information in deep neural networks},
  author={Buhrmester, Vanessa and M{\"u}nch, David and Bulatov, Dimitri and Arens, Michael},
  booktitle={Iberian conference on pattern recognition and image analysis},
  pages={302--316},
  year={2019},
  organization={Springer}
}

@inproceedings{de2021impact,
  title={Impact of colour on robustness of deep neural networks},
  author={De, Kanjar and Pedersen, Marius},
  booktitle={Proceedings of the IEEE/CVF international conference on computer vision},
  pages={21--30},
  year={2021}
}

@inproceedings{hendrycks2021many,
  title={The many faces of robustness: A critical analysis of out-of-distribution generalization},
  author={Hendrycks, Dan and Basart, Steven and Mu, Norman and Kadavath, Saurav and Wang, Frank and Dorundo, Evan and Desai, Rahul and Zhu, Tyler and Parajuli, Samyak and Guo, Mike and others},
  booktitle={Proceedings of the IEEE/CVF international conference on computer vision},
  pages={8340--8349},
  year={2021}
}

@article{herbreteau2023normalization,
  title={Normalization-equivariant neural networks with application to image denoising},
  author={Herbreteau, S{\'e}bastien and Moebel, Emmanuel and Kervrann, Charles},
  journal={Advances in Neural Information Processing Systems},
  volume={36},
  pages={5706--5728},
  year={2023}
}

@article{ho2020denoising,
  title={Denoising diffusion probabilistic models},
  author={Ho, Jonathan and Jain, Ajay and Abbeel, Pieter},
  journal={Advances in neural information processing systems},
  volume={33},
  pages={6840--6851},
  year={2020}
}

@inproceedings{xie2024decomposing,
  title={Decomposing the generalization gap in imitation learning for visual robotic manipulation},
  author={Xie, Annie and Lee, Lisa and Xiao, Ted and Finn, Chelsea},
  booktitle={2024 IEEE International Conference on Robotics and Automation (ICRA)},
  pages={3153--3160},
  year={2024},
  organization={IEEE}
}

@article{pumacay2024colosseum,
  title={The colosseum: A benchmark for evaluating generalization for robotic manipulation},
  author={Pumacay, Wilbert and Singh, Ishika and Duan, Jiafei and Krishna, Ranjay and Thomason, Jesse and Fox, Dieter},
  journal={arXiv preprint arXiv:2402.08191},
  year={2024}
}

@article{fei2025libero,
  title={Libero-plus: In-depth robustness analysis of vision-language-action models},
  author={Fei, Senyu and Wang, Siyin and Shi, Junhao and Dai, Zihao and Cai, Jikun and Qian, Pengfang and Ji, Li and He, Xinzhe and Zhang, Shiduo and Fei, Zhaoye and others},
  journal={arXiv preprint arXiv:2510.13626},
  year={2025}
}

@inproceedings{tobin2017domain,
  title={Domain randomization for transferring deep neural networks from simulation to the real world},
  author={Tobin, Josh and Fong, Rachel and Ray, Alex and Schneider, Jonas and Zaremba, Wojciech and Abbeel, Pieter},
  booktitle={2017 IEEE/RSJ international conference on intelligent robots and systems (IROS)},
  pages={23--30},
  year={2017},
  organization={IEEE}
}

@article{chen2023genaug,
  title={Genaug: Retargeting behaviors to unseen situations via generative augmentation},
  author={Chen, Zoey and Kiami, Sho and Gupta, Abhishek and Kumar, Vikash},
  journal={arXiv preprint arXiv:2302.06671},
  year={2023}
}

@book{gallier2020differential,
  title={Differential geometry and lie groups},
  author={Gallier, Jean Quaintance and Quaintance, Jocelyn},
  volume={12},
  year={2020},
  publisher={Springer}
}

@book{rotman2012introduction,
  title={An introduction to the theory of groups},
  author={Rotman, Joseph J},
  year={2012},
  publisher={Springer Science \& Business Media}
}

@book{fulton2013representation,
  title={Representation theory: a first course},
  author={Fulton, William and Harris, Joe},
  year={2013},
  publisher={Springer Science \& Business Media}
}

@book{serre1977linear,
  title={Linear representations of finite groups},
  author={Serre, Jean-Pierre and others},
  volume={42},
  year={1977},
  publisher={Springer}
}

@book{olver1999classical,
  title={Classical invariant theory},
  author={Olver, Peter J},
  number={44},
  year={1999},
  publisher={Cambridge University Press}
}

@book{humphreys1996course,
  title={A course in group theory},
  author={Humphreys, John F},
  volume={6},
  year={1996},
  publisher={OUP Oxford}
}

@article{zini2022planckian,
  title={Planckian jitter: countering the color-crippling effects of color jitter on self-supervised training},
  author={Zini, Simone and Gomez-Villa, Alex and Buzzelli, Marco and Twardowski, Bart{\'L} and Bagdanov, Andrew D and Van de Weijer, Joost and others},
  journal={arXiv preprint arXiv:2202.07993},
  year={2022}
}

@inproceedings{guo2020zero,
  title={Zero-reference deep curve estimation for low-light image enhancement},
  author={Guo, Chunle and Li, Chongyi and Guo, Jichang and Loy, Chen Change and Hou, Junhui and Kwong, Sam and Cong, Runmin},
  booktitle={Proceedings of the IEEE/CVF conference on computer vision and pattern recognition},
  pages={1780--1789},
  year={2020}
}

@article{wei2018deep,
  title={Deep retinex decomposition for low-light enhancement},
  author={Wei, Chen and Wang, Wenjing and Yang, Wenhan and Liu, Jiaying},
  journal={arXiv preprint arXiv:1808.04560},
  year={2018}
}

@inproceedings{finn2016deep,
  title={Deep spatial autoencoders for visuomotor learning},
  author={Finn, Chelsea and Tan, Xin Yu and Duan, Yan and Darrell, Trevor and Levine, Sergey and Abbeel, Pieter},
  booktitle={2016 IEEE International Conference on Robotics and Automation (ICRA)},
  pages={512--519},
  year={2016},
  organization={IEEE}
}

@inproceedings{wu2018group,
  title={Group normalization},
  author={Wu, Yuxin and He, Kaiming},
  booktitle={Proceedings of the European conference on computer vision (ECCV)},
  pages={3--19},
  year={2018}
}

@article{jin2025physically,
  title={Physically-based Lighting Generation for Robotic Manipulation},
  author={Jin, Shutong and Wang, Lezhong and Temming, Ben and Pokorny, Florian T},
  journal={arXiv preprint arXiv:2508.01442},
  year={2025}
}

@inproceedings{tie2025seed,
  title={Et-seed: Efficient trajectory-level se (3) equivariant diffusion policy},
  author={Tie, Chenrui and Chen, Yue and Wu, Ruihai and Dong, Boxuan and Li, Zeyi and Gao, Chongkai and Dong, Hao},
  booktitle={International Conference on Learning Representations},
  volume={2025},
  pages={60114--60132},
  year={2025}
}

@inproceedings{messaoudi2026designing,
  title={Designing Affine-Invariant Neural Networks for Photometric Corruption Robustness and Generalization},
  author={Messaoudi, Mounir and Rapilly, Quentin and Herbreteau, S{\'e}bastien and Badoual, Ana{\"\i}s and Kervrann, Charles},
  booktitle={International Conference on Learning Representations},
  volume={2026},
  year={2026}
}

@inproceedings{baek2024unexplored,
  title={Unexplored faces of robustness and out-of-distribution: Covariate shifts in environment and sensor domains},
  author={Baek, Eunsu and Park, Keondo and Kim, Jiyoon and Kim, Hyung-Sin},
  booktitle={Proceedings of the IEEE/CVF Conference on Computer Vision and Pattern Recognition},
  pages={22294--22303},
  year={2024}
}

@article{zhu2025se,
  title={Se (3)-equivariant diffusion policy in spherical fourier space},
  author={Zhu, Xupeng and Wang, Fan and Walters, Robin and Shi, Jane},
  journal={arXiv preprint arXiv:2507.01723},
  year={2025}
}

@article{niu2023vergnet,
  title={Vergnet: Visual enhancement guided robotic grasp detection under low-light condition},
  author={Niu, Mingdi and Lu, Zhenyu and Chen, Lu and Yang, Jing and Yang, Chenguang},
  journal={IEEE Robotics and Automation Letters},
  volume={8},
  number={12},
  pages={8541--8548},
  year={2023},
  publisher={IEEE}
}

@article{sun2026light,
  title={Light-resilient visual regression of strawberry ripeness for robotic harvesting},
  author={Sun, Meili and Hu, Shimin and Zhao, Chunjiang and Xiong, Ya},
  journal={Computers and Electronics in Agriculture},
  volume={241},
  pages={111169},
  year={2026},
  publisher={Elsevier}
}

@inproceedings{todorov2012mujoco,
  title={Mujoco: A physics engine for model-based control},
  author={Todorov, Emanuel and Erez, Tom and Tassa, Yuval},
  booktitle={2012 IEEE/RSJ international conference on intelligent robots and systems},
  pages={5026--5033},
  year={2012},
  organization={IEEE}
}

@article{song2020denoising,
  title={Denoising diffusion implicit models},
  author={Song, Jiaming and Meng, Chenlin and Ermon, Stefano},
  journal={arXiv preprint arXiv:2010.02502},
  year={2020}
}

@inproceedings{rombach2022high,
  title={High-resolution image synthesis with latent diffusion models},
  author={Rombach, Robin and Blattmann, Andreas and Lorenz, Dominik and Esser, Patrick and Ommer, Bj{\"o}rn},
  booktitle={Proceedings of the IEEE/CVF conference on computer vision and pattern recognition},
  pages={10684--10695},
  year={2022}
}

@inproceedings{he2016deep,
  title={Deep residual learning for image recognition},
  author={He, Kaiming and Zhang, Xiangyu and Ren, Shaoqing and Sun, Jian},
  booktitle={Proceedings of the IEEE conference on computer vision and pattern recognition},
  pages={770--778},
  year={2016}
}

@article{lim2017enhanced,
  title={Enhanced deep residual networks for single image super-resolution},
  author={Lim, Bee and Son, Sanghyun and Kim, Heewon and Nah, Seungjun and Lee, Kyoung Mu},
  journal={arXiv preprint arXiv:1707.02921},
  year={2017}
}

@article{theis2017lossy,
  title={Lossy image compression with compressive autoencoders},
  author={Theis, Lucas and Shi, Wenzhe and Cunningham, Andrew and Husz{\'a}r, Ferenc},
  journal={arXiv preprint arXiv:1703.00395},
  year={2017}
}

@inproceedings{perez2018film,
  title={Film: Visual reasoning with a general conditioning layer},
  author={Perez, Ethan and Strub, Florian and De Vries, Harm and Dumoulin, Vincent and Courville, Aaron},
  booktitle={Proceedings of the AAAI conference on artificial intelligence},
  volume={32},
  number={1},
  year={2018}
}

@inproceedings{devlin2019bert,
  title={Bert: Pre-training of deep bidirectional transformers for language understanding},
  author={Devlin, Jacob and Chang, Ming-Wei and Lee, Kenton and Toutanova, Kristina},
  booktitle={Proceedings of the 2019 conference of the North American chapter of the association for computational linguistics: human language technologies, volume 1 (long and short papers)},
  pages={4171--4186},
  year={2019}
}

@article{healey1994radiometric,
  title={Radiometric CCD camera calibration and noise estimation},
  author={Healey, Glenn E and Kondepudy, Raghava},
  journal={IEEE Transactions on Pattern Analysis and Machine Intelligence},
  volume={16},
  number={3},
  pages={267--276},
  year={1994},
  publisher={IEEE}
}

@article{foi2008practical,
  title={Practical Poissonian-Gaussian noise modeling and fitting for single-image raw-data},
  author={Foi, Alessandro and Trimeche, Mejdi and Katkovnik, Vladimir and Egiazarian, Karen},
  journal={IEEE transactions on image processing},
  volume={17},
  number={10},
  pages={1737--1754},
  year={2008},
  publisher={IEEE}
}

@incollection{debevec2023recovering,
  title={Recovering high dynamic range radiance maps from photographs},
  author={Debevec, Paul E and Malik, Jitendra},
  booktitle={Seminal Graphics Papers: Pushing the Boundaries, Volume 2},
  pages={643--652},
  year={2023}
}

@article{grossberg2004modeling,
  title={Modeling the space of camera response functions},
  author={Grossberg, Michael D and Nayar, Shree K},
  journal={IEEE transactions on pattern analysis and machine intelligence},
  volume={26},
  number={10},
  pages={1272--1282},
  year={2004},
  publisher={IEEE}
}

@article{exel1998partial,
  title={Partial actions of groups and actions of inverse semigroups},
  author={Exel, Ruy},
  journal={Proceedings of the American Mathematical Society},
  volume={126},
  number={12},
  pages={3481--3494},
  year={1998}
}

@article{zhang2026rohil,
  title={RoHIL: Robust Human-in-the-Loop Robotic Reinforcement Learning Against Illumination Variations},
  author={Zhang, Shuoqin and Xiong, Yixin and Gao, Xiru and Liu, Kai and Wang, Ke and Zhou, Xichuan and Hu, Zhe},
  journal={arXiv preprint arXiv:2605.19924},
  year={2026}
}

@article{chisari2024learning,
  title={Learning robotic manipulation policies from point clouds with conditional flow matching},
  author={Chisari, Eugenio and Heppert, Nick and Argus, Max and Welschehold, Tim and Brox, Thomas and Valada, Abhinav},
  journal={arXiv preprint arXiv:2409.07343},
  year={2024}
}

@article{qi2017pointnet++,
  title={Pointnet++: Deep hierarchical feature learning on point sets in a metric space},
  author={Qi, Charles Ruizhongtai and Yi, Li and Su, Hao and Guibas, Leonidas},
  journal={Advances in neural information processing systems},
  volume={30},
  year={2017}
}

@article{zhu2022viola,
  title={VIOLA: Imitation Learning for Vision-Based Manipulation with Object Proposal Priors},
  author={Zhu, Yifeng and Joshi, Abhishek and Stone, Peter and Zhu, Yuke},
  journal={arXiv preprint arXiv:2210.11339},
  doi={10.48550/arXiv.2210.11339},
  year={2022}
}

@article{nakanishi2008operational,
  title={Operational space control: A theoretical and empirical comparison},
  author={Nakanishi, Jun and Cory, Rick and Mistry, Michael and Peters, Jan and Schaal, Stefan},
  journal={The International Journal of Robotics Research},
  volume={27},
  number={6},
  pages={737--757},
  year={2008},
  publisher={SAGE Publications, Sage UK: London, England}
}

@inproceedings{zhou2019continuity,
  title={On the continuity of rotation representations in neural networks},
  author={Zhou, Yi and Barnes, Connelly and Lu, Jingwan and Yang, Jimei and Li, Hao},
  booktitle={2019 IEEE/CVF Conference on Computer Vision and Pattern Recognition (CVPR)},
  pages={5738--5746},
  year={2019},
  organization={IEEE}
}
\clearpage
\appendix
\section{Method}
This section details the implementation of the method and the comparisons evaluated in the main paper, together with the measure used to quantify invariance.
The CARROT encoder is described through its group-equivariant backbone, the lifting of each observation, and the keypoint readout that conditions the denoiser.
The baselines are organized by where they enter the policy pipeline, as train-time augmentation, train-time observation preprocessing, or evaluation-time restoration.
An orbit distance measures how far a perturbed observation lies from the lifted orbit of its clean counterpart and bounds the resulting change in the encoder output.
The alternative invariance mechanisms, namely shared convolutions, on-orbit augmentation, and test-time ensembling, differ from CARROT in how the encoder treats the luminance orbit.
\subsection{CARROT Implementation Details}
\paragraph{Backbone.} The encoder is a ResNet18~\citep{he2016deep} in which every convolution and normalization layer, including those of the input block, is replaced by its group-equivariant counterpart. The initial $7\times7$ strided convolution is therefore a group convolution over the lifted orbit, and the max pool that follows acts only on the spatial grid. Each normalization layer is the group equivariant group normalization of Section~\ref{subsec:method_instance_norm}, which extends the group normalization~\citep{wu2018group} used by Diffusion Policy~\citep{chi2025diffusion} by sharing statistics across the luminance orbit. The $1/\sqrt{R}$ channel width reduction is applied at every group convolution, so the parameter count of CARROT matches that of Diffusion Policy in Table~\ref{tab:latency}. The luminance subgroup is obtained from the hypertoroidal construction of~\citet{yang2026hypertoroidal} by setting the orders of the hue and saturation factors to one. Since the third-person and wrist cameras are positioned differently, the same change in illumination shifts their luminance by different amounts (Table~\ref{tab:realworld_dark}), so each camera has its own encoder without weight sharing, which lifts and pools its observations independently.
\paragraph{Lifting.} The lift is computed in the forward pass, so the $R$ copies of each observation are never stored as an expanded dataset. Each image is converted from the symmetric range used by the baseline to HSL, shifted, and converted back before ImageNet normalization, so the group action acts on displayable values and every element of the orbit is normalized identically.
\paragraph{Readout.} Before the spatial softmax, a $1\times1$ convolution reduces the pooled feature map to the $K_{\mathrm{kp}}$ keypoint channels. Since it acts on an already invariant map, it preserves the invariance of the encoder while letting the number of keypoints be chosen independently of the trunk width. The concatenated keypoints have the same per-camera dimension $2K_{\mathrm{kp}}$ as the spatial softmax readout of the baseline encoder, so they enter the FiLM conditioning~\citep{perez2018film} of the denoiser exactly as in the baseline. No other component is retuned when the encoder is replaced, so the differences from Diffusion Policy in Tables~\ref{tab:libero_vanilla} to~\ref{tab:libero_plus_language} are attributable to the encoder.
\subsection{Baseline Implementation Details}\label{apx:baselines}
The baselines are grouped by where they enter the policy pipeline, through the training data, the observations at evaluation, or the input modality.
Every image-based policy shares the diffusion head, training recipe, and hyperparameters of Diffusion Policy~\citep{chi2025diffusion}, listed in Appendix~\ref{apx:training}.
\subsubsection{Augmentation}
DP-Jitter, DP-Planck, and DP-DeepAug are precomputed when the demonstrations are converted to the training cache, and each policy is trained with exactly one of them.
Each demonstration receives one fixed perturbation, drawn from a generator seeded by its index and applied to every frame and both cameras.
The training set contains both the clean and the perturbed copy of every demonstration.
All three transforms operate on 8-bit images and clip rather than rescale.
\paragraph{DP-Jitter.} DP-Jitter applies a lightness-only additive shift $\left(0, 0, \delta_{l}\right)$ in HSL with $\delta_{l}\sim\mathcal{U}\left(-0.5, 0.5\right)$, clipping the lightness channel to $\left[0, 1\right]$.
\paragraph{DP-Planck.} DP-Planck~\citep{zini2022planckian} shifts the white balance along the blackbody locus by scaling the red and blue channels with ratios drawn uniformly from the 25 entries of a Planckian locus table, leaving green unchanged, which makes it a chromatic perturbation.
\paragraph{DP-DeepAug.} DP-DeepAug adapts DeepAugment~\citep{hendrycks2021many}, which passes images through an EDSR~\citep{lim2017enhanced} or CAE~\citep{theis2017lossy} image-to-image network whose weights and activations are distorted by randomly sampled operations. The original trains on the clean data together with one distorted copy from each network, whereas DP-DeepAug follows the protocol shared by the augmentations above and adds a single copy of each demonstration distorted by one network, drawn with equal probability. The corruption options are drawn uniformly from the ranges of the original implementation, and DP-DeepAug exposes no severity parameter. Only the distorting network is fixed for each demonstration, so its output still varies from frame to frame with the frame content.
\subsubsection{Observation Preprocessing}
\paragraph{Stem-OB.} Stem-OB~\citep{hu2025stem} replaces every training frame of both cameras with its diffusion-inverted counterpart, while evaluation reads raw frames, so the only difference from Diffusion Policy is the pixels on which the encoder is fit. Inversion uses a frozen Stable Diffusion 2.1 checkpoint~\citep{rombach2022high} with the default recipe of the reference implementation, in which error reduction and the reverse pass are disabled. Each frame therefore passes through the VAE encoder, a closed-form forward noising step, and the VAE decoder without invoking the image model's own denoising network, so the stem is obtained by forward noising rather than by reverse-pass DDIM~\citep{song2020denoising} inversion, although both are referred to as inversion in the literature. Noising stops at the tenth index of a 50-step schedule, which corresponds to $t\approx181$ on the 1000-step noise schedule. We selected this strength over 5 steps after a probe in which the two tied, resolving the tie toward the stronger setting by a rule fixed before the probe. Inversion runs at $768\times768$ with nearest-neighbor resizing in and out and is applied to every training frame.
\subsubsection{Restoration}
RetinexNet and Zero-DCE are pretrained and frozen, and each is inserted at evaluation as a per-camera restoration step in front of the vanilla Diffusion Policy checkpoints with EMA weights, so restoration changes only the observations that a fixed policy sees.
Restoration runs before observation normalization at the native observation resolution, $128\times128$ in simulation, and its output is rounded and clipped to 8 bits.
For luminance-shifted LIBERO, the shift is applied before restoration, matching the deployment order in which a restoration module receives an already degraded frame.
\paragraph{RetinexNet.} RetinexNet~\citep{wei2018deep} uses the published decomposition and relighting sub-networks, retrieved from \url{github.com/weichen582/RetinexNet}, and reconstructs each frame as the product of the estimated reflectance and the adjusted illumination map.
\paragraph{Zero-DCE.} Zero-DCE~\citep{guo2020zero} uses the published checkpoint of roughly 79k parameters, retrieved from \url{github.com/Li-Chongyi/Zero-DCE}, and applies its eight-iteration enhancement curve $x\leftarrow x+A\left(x^{2}-x\right)$ with pixel-wise curve parameters $A$ predicted by the network.
\subsubsection{Point-Cloud Policies}
DP3 and DP3-RGB are evaluated only on the real robot and are trained on the same 40 demonstrations per task as the image-based policies.
\paragraph{DP3.} DP3~\citep{ze20243d} conditions the denoiser on a point cloud rather than on RGB images, so its observation carries scene geometry without appearance. Following the original implementation, DP3 observes the scene through the third-person camera alone. At every step, its depth image is back-projected into a point cloud, cropped to the workspace, and downsampled by farthest point sampling to the 1024 points used in the real-world tasks of~\citet{ze20243d}, and the point coordinates are encoded by the DP3 point encoder. This construction runs for every observation at deployment and is included in the total latency of Table~\ref{tab:latency}.
\paragraph{DP3-RGB.} DP3-RGB is the colored-point variant ablated by~\citet{ze20243d}, which appends the RGB value of each point to its coordinates. It is otherwise identical to DP3, so it recovers chromatic information together with sensitivity to illumination.
\subsection{Exact Invariance}\label{apx:proofs}
This section proves that a modeled luminance shift cyclically permutes the lifted representation, the property from which the invariance of the encoder follows.
For an image in HSL color space $x=\left(x_{h},x_{s},x_{l}\right)$ with lower branch coordinate $\theta_{l}=\pi x_{l}$, the action of an element $l_{i}$ of the luminance group $L_{R}=\left\{l_{i}=\nicefrac{2\pi i}{R}\mid i=0,\ldots,R-1\right\}$ is
\begin{equation}
    \varphi_{l}\left(l_{i},x\right) = \left(x_{h},\; x_{s},\; \mathcal{C}\left(\left(\theta_{l}+l_{i}\right)\bmod 2\pi\right)\right),
\end{equation}
where the covering map $\mathcal{C}$ equals $\theta/\pi$ on the lower branch $\left[0,\pi\right]$ and $2-\theta/\pi$ on the upper branch $\left[\pi,2\pi\right]$.
The lift is $f^{0}\left(l_{i}\right)=\varphi_{l}\left(l_{i},x\right)$, and $\phi_{l}\left(l_{j},\cdot\right)$ denotes the cyclic permutation that replaces the entry at each $l_{i}$ with the entry at $l_{i}\cdot l_{j}$.
A modeled luminance shift by $l_{j}$ is the uniform shift $x_{l}\mapsto x_{l}+\delta_{j}$ of an image whose pixels all satisfy $x_{l}+\delta_{j}\in\left[0,1\right]$, with $\delta_{j}\in\left(-1,1\right]$ congruent to $2j/R$ modulo $2$.
\begin{proposition}\label{prop:lift}
    Let $x'$ be a modeled luminance shift of $x$ by $l_{j}$.
    Then $f^{0}\left(x'\right)=\phi_{l}\left(l_{j},f^{0}\left(x\right)\right)$.
\end{proposition}
\begin{proof}
    At every pixel, $\pi\left(x_{l}+\delta_{j}\right)$ lies in $\left[0,\pi\right]$ and is congruent to $\theta_{l}+l_{j}$ modulo $2\pi$, so the rotated coordinate $\left(\theta_{l}+l_{j}\right)\bmod 2\pi$ stays on the lower branch.
    The action of $l_{j}$ therefore returns the luminance $x_{l}+\delta_{j}$, so $x'=\varphi_{l}\left(l_{j},x\right)$, and the lower branch coordinate of $x'$ is the rotated coordinate.
    Acting on $x'$ with $l_{i}$ then gives the luminance $\mathcal{C}\left(\left(\theta_{l}+l_{j}+l_{i}\right)\bmod 2\pi\right)$, which is that of $\varphi_{l}\left(l_{i}\cdot l_{j},x\right)$, while hue and saturation are unchanged.
    Hence $f^{0}\left(x'\right)\left(l_{i}\right)=f^{0}\left(x\right)\left(l_{i}\cdot l_{j}\right)$ for every $l_{i}\in L_{R}$.
\end{proof}
The argument is pixelwise, so under a clipped shift the lift departs from the permuted lift only at the clipped pixels.
Group convolutions commute with this permutation~\citep{cohen2016group}, the group equivariant group normalization computes its statistics over the complete orbit, and the remaining layers act identically at every element of $L_{R}$.
The permutation therefore reaches the final feature map, where orbit max pooling removes it, which gives $E_{\psi}\left(x'\right)=E_{\psi}\left(x\right)$.
\subsection{Quantifying Invariance}\label{apx:quant_invariance}
The encoder is exactly invariant to modeled luminance shifts, but illumination changes encountered in practice may clip pixels, vary across the image, or fall between elements of $L_{R}$.
This section measures how far such a perturbation departs from the modeled action and bounds the resulting change in the encoder output. 
To this end, we write the encoder as $E_{\psi}=\Psi\circ F^{0}$
with
\begin{equation}
    F^{0}\left(x\right)=\left(f^{0}\left(l_{0}\right),\ldots,f^{0}\left(l_{R-1}\right)\right),
\end{equation}
which is the lifted representation of an observation $x$ and $\Psi$ comprises the remaining layers.
The group convolutions and normalization commute with the cyclic permutation $\phi_{l}$, and orbit pooling is invariant to it, so $\Psi\left(\phi_{l}\left(l_{j},F\right)\right)=\Psi\left(F\right)$ for every input $F$, whether or not $F$ is the lift of a transformed observation.
For a clean observation $x$ and a perturbed observation $x'$, we measure the departure from the modeled action with the orbit distance
\begin{equation}
    \Delta\left(x',x\right)=\min_{l_{j}\in L_{R}}\left\lVert F^{0}\left(x'\right)-\phi_{l}\left(l_{j},F^{0}\left(x\right)\right)\right\rVert,
\end{equation}
which is the distance from the lift of $x'$ to the closest element of the lifted orbit of $x$.
Since $\Psi$ takes the same value on all element on orbit, the change in the encoder output satisfies
\begin{equation}
    \left\lVert E_{\psi}\left(x'\right)-E_{\psi}\left(x\right)\right\rVert\le L_{\Psi}\,\Delta\left(x',x\right),
\end{equation}
where $L_{\Psi}$ is a Lipschitz constant of $\Psi$, which is finite since every layer of $\Psi$ is Lipschitz.
The orbit distance vanishes for a modeled luminance shift, recovering exact invariance.
For a uniform shift between group elements without clipping, it is bounded by a multiple of the luminance offset from the nearest element on the circle, which is at most $\nicefrac{1}{R}$.
The covering map has slope $\pm1/\pi$, so each luminance entry of $F^{0}\left(x'\right)$ lies within this offset of the matching entry of the nearest element of the lifted orbit.

We report $\Delta$ in two forms.
The unabsorbed fraction
\begin{equation}
    \eta = \frac{\Delta}{\left\lVert F^{0}\left(x'\right)-F^{0}\left(x\right)\right\rVert},
\end{equation}
divides the orbit distance by the distance obtained without group correction and measures the share of the lifted change that no element of $L_{R}$ explains.
$\eta$ ranges from 0, when a group element explains the change exactly, to 1, when none improves on the identity.
The relative orbit distance
\begin{equation}
    \xi = \frac{\Delta}{\left\lVert F^{0}\left(x\right)\right\rVert},
\end{equation}
measures the size of this unexplained part relative to the lifted observation itself.
\subsection{Ablation Implementation Details}\label{apx:ablation}
The three variants in Table~\ref{tab:libero_plus_ablation} differ from CARROT in how the vision encoder treats the luminance orbit of each observation, with $R=8$ throughout.
\paragraph{Shared-Conv.} 
Every group convolution of CARROT is replaced by a conventional convolution applied separately to each of the eight lifted copies, so no information is exchanged between elements of the orbit.
The lifting layer, the group equivariant group normalization, the orbit max pooling ahead of the spatial softmax, and the keypoint readout are unchanged.
Like CARROT, the variant is therefore exactly invariant to modeled luminance shifts. 
Its vision encoder requires the same number of convolution multiply-accumulate operations as that of CARROT and has a comparable parameter count of 22.4M against 22.3M.
\paragraph{On-Orbit Augmentation.} 
A standard Diffusion Policy is trained on observations transformed by the same group on which CARROT is built.
During training, each camera image is transformed by all eight elements of $L_{8}$, with each element applied to both frames of the observation horizon.
The policy therefore sees the complete orbit of every observation, whereas DP-Jitter in Appendix~\ref{apx:baselines} applies a single luminance shift drawn once per demonstration.
Observations are left unmodified at evaluation.
\paragraph{Test-Time Ensembling.}
Orbit pooling is added at evaluation to the unaugmented Diffusion Policy. 
The eight orbit copies of each image pass through the encoder and are max pooled ahead of the spatial softmax, at the same point at which CARROT pools. 
Pooling is computed independently for each image, so no statistics are shared across evaluation environments. 
The variant therefore isolates the contribution of orbit pooling alone.
\section{Dataset and Hardware Setup}
\begin{figure}[t]
    \centering
    \begin{subfigure}[t]{\columnwidth}
        \centering
        \includegraphics[width=\columnwidth]{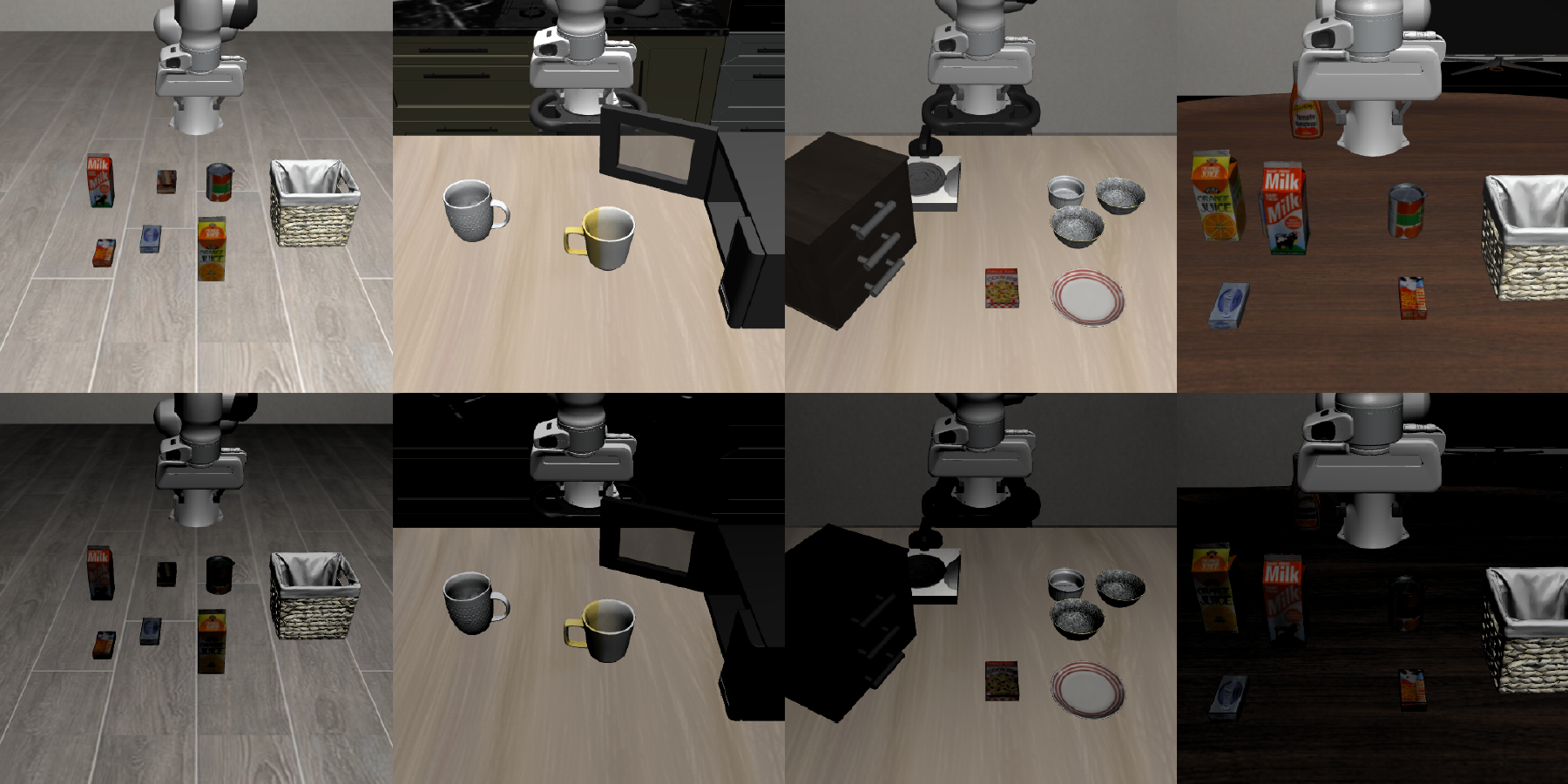}
        \caption{\textbf{Luminance Shifted LIBERO.} We show \textbf{(top)} original observations and \textbf{(bottom)} their counterparts under the clipped luminance shift of Equation~\eqref{eqn:libero_saturation_shift}.}
        \label{fig:T3DP_LIBEROSaturationDatasetDemo}
    \end{subfigure}
    \vskip 2mm
    \begin{subfigure}[t]{\columnwidth}
        \centering
        \includegraphics[width=\columnwidth]{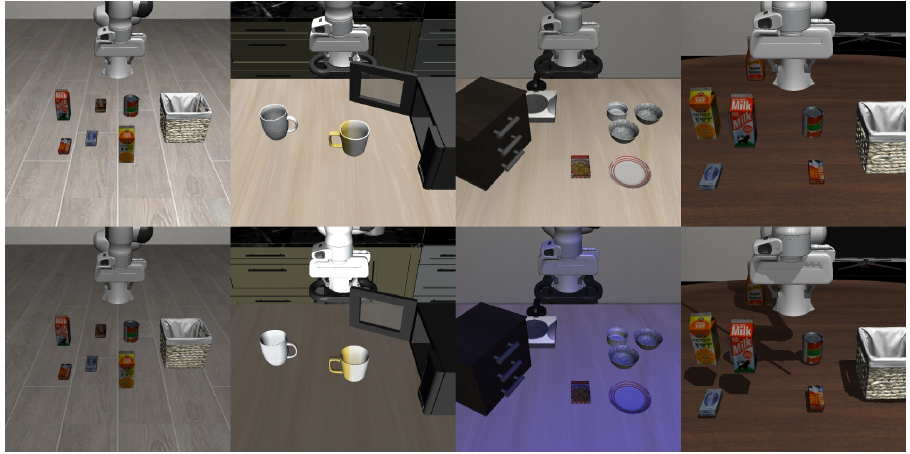}
        \caption{\textbf{LIBERO-Plus Lighting Perturbations.} We show \textbf{(top)} original observations and \textbf{(bottom)} their counterparts under perturbed lighting, varying intensity, direction, hue, and shadow casting from left to right.}
        \label{fig:T3DP_LIBEROPlusDatasetDemo}
    \end{subfigure}
    \caption{\textbf{LIBERO and LIBERO-Plus Example Images.} We show clean and perturbed observations from \textbf{(a)} luminance shifted LIBERO and \textbf{(b)} LIBERO-Plus lighting.}
    \label{fig:T3DP_LIBERODatasetDemo}
\end{figure}

This section describes the simulated benchmarks and the physical platform on which the policies of the main paper are trained and evaluated.
Both simulated benchmarks share the Goal, Object, and Spatial suites and demonstrations of LIBERO~\citep{liu2023libero} and differ in how illumination is perturbed at evaluation.
Luminance shifted LIBERO darkens the rendered observations directly, whereas LIBERO-Plus~\citep{fei2025libero} edits the scene lighting before rendering.
Figure~\ref{fig:T3DP_LIBERODatasetDemo} contrasts clean and perturbed observations from both benchmarks.
The real world setup carries the evaluation onto a physical robot, where the lighting change originates from the light sources around the workspace and reaches the policy through the camera response rather than as an edit to the observation.
\subsection{LIBERO}
LIBERO~\citep{liu2023libero} is a simulation benchmark for language conditioned manipulation built in MuJoCo~\citep{todorov2012mujoco}, in which a Franka Emika Panda arm performs tabletop tasks in kitchen and living room scenes.
An observation pairs a third person and a wrist RGB image, each rendered at $128\times128$, with the low dimensional robot state read from the simulator, and actions are delta end effector poses with a gripper command.
We use the Goal, Object, and Spatial suites, each containing ten tasks with 50 human demonstrations per task.
The three suites differ in what varies across their tasks, with Spatial varying object layout for a fixed object set, Object varying the object set for a fixed layout, and Goal varying the instructed goal while objects and layout are held fixed.
\paragraph{Luminance Shift.}
We evaluate under a global luminance shift applied to the rendered RGB observations.
Writing a pixel in HSL coordinates as $x=\left(x_{h},x_{s},x_{l}\right)$, the perturbed pixel is
\begin{equation}
    x'= \left(x_{h},x_{s},\mathrm{max}\left(x_{l}-0.25,0.0\right)\right).\label{eqn:apx_libero_luminance_shift}
\end{equation}
Hue and saturation are therefore preserved, while lightness is lowered uniformly by 0.25 and clipped at zero.
For pixels with lightness of at least 0.25, the shift coincides with an element of the discretized luminance group used by CARROT.
Clipping maps lightness below 0.25 to zero, which is not invertible and breaks the exact group action on those pixels, so we treat this evaluation as a near orbit shift rather than an exact test of the invariance guarantee.
The same shift is applied at evaluation to every rendered frame of both cameras, while the task, scene, success criteria, and robot state are left unchanged, so the observations are the only source of distribution shift.
Policies are trained on the unperturbed demonstrations and evaluated without adaptation.
Under both the unperturbed and the shifted observations, each policy is rolled out from the 50 canonical test initial states of its task, and success rates are reported as the mean and standard deviation over three training seeds.
\subsection{LIBERO-Plus}
LIBERO-Plus~\citep{fei2025libero} is a robustness benchmark that perturbs the LIBERO tasks along seven dimensions, namely object layout, camera viewpoint, robot initial state, language instruction, light conditions, background texture, and sensor noise, yielding 10,030 task variants.
We use only the light conditions dimension, on the same Goal, Object, and Spatial suites, for which the benchmark provides 279, 297, and 292 variants, 868 in total.
Every light variant is generated by editing the scene definition of an existing LIBERO task, so the observation modality, action space, demonstration format, and success criteria are inherited unchanged and policies trained on the original demonstrations can be evaluated directly.
Each single task policy is evaluated without adaptation on every light variant of its task, starting from the robot initial states of the training demonstrations, and success rates are reported as the mean and standard deviation over three training seeds.
\paragraph{Lighting Factors.}
The light dimension comprises four sub dimensions, each editing one attribute of the scene light sources.
Diffuse sets the RGB color of the light reflected by object surfaces, so a uniform scaling produces a spatially near uniform change in image luminance, while unequal channel gains recolor the scene as a white balance shift would.
Direction reorients the directional source, relocating shading gradients and specular highlights even when mean brightness is approximately preserved.
Specular sets the strength of the highlights reflected by glossy surfaces.
Shadow toggles whether the arm and objects cast shadows, removing scene radiance in regions that no per pixel transformation can recover.
A single variant may alter several of these attributes at once.
All four act at the renderer level and therefore perturb the third person and wrist cameras simultaneously, while the low dimensional robot state is preserved exactly, so the observations remain the only source of distribution shift.
\subsection{Real World Setup}
\begin{figure}[t]
    \centering
    \includegraphics[width=\linewidth]{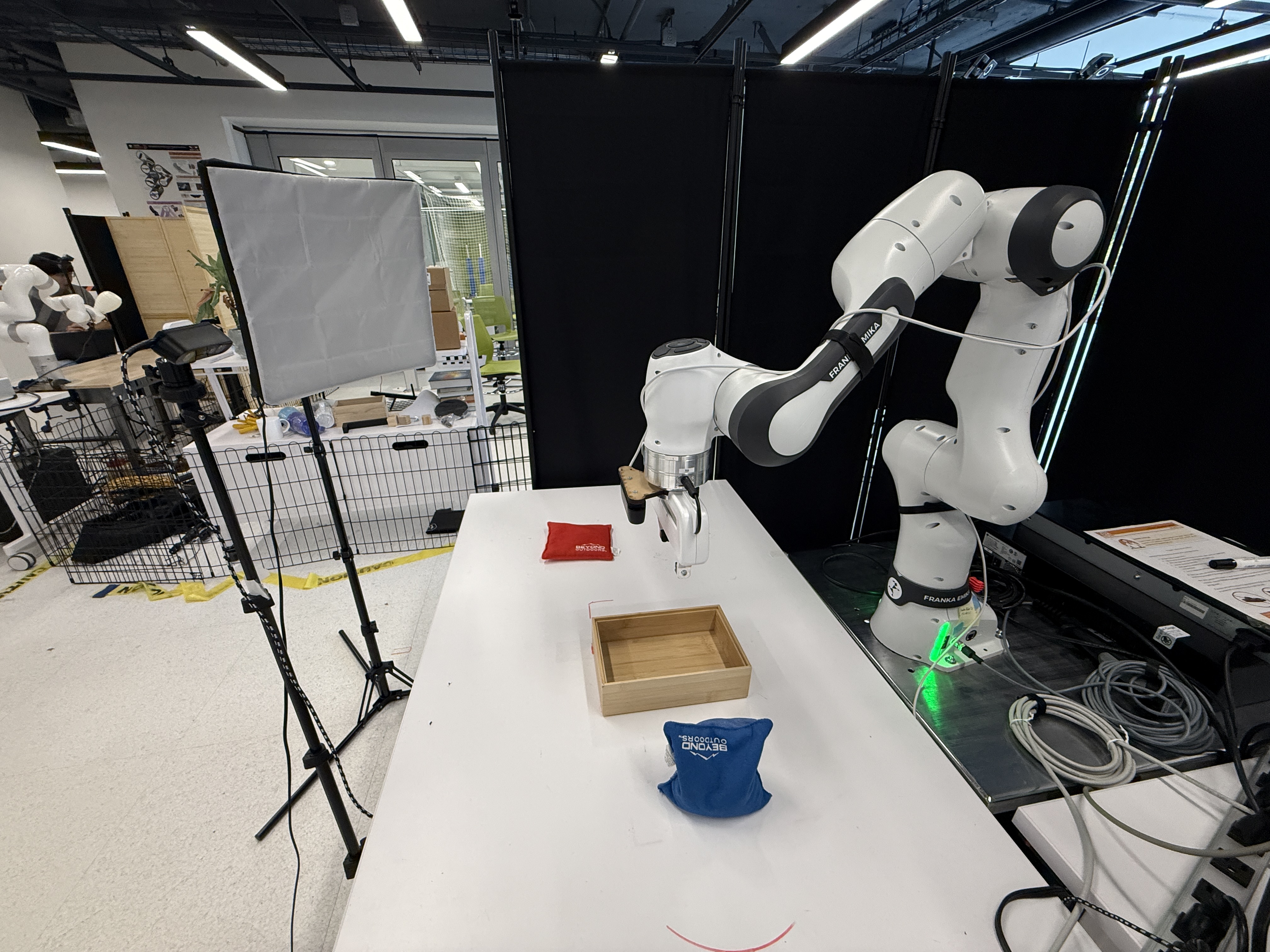}
    \caption{\textbf{Real World Setup.} We show the Franka FR3 with the Bean task on the table, the third person camera on its tripod, and one of the two studio lights that produce the reduced lighting.}
    \label{fig:CARROT_Hardware}
\end{figure}

\paragraph{Hardware.}
The real world experiments use a Franka FR3 arm equipped with a Franka Hand (shown in Figure~\ref{fig:CARROT_Hardware}) and driven by the operational space pose controller~\citep{nakanishi2008operational} of Deoxys~\citep{zhu2022viola} with a proportional gain of $k_{p}=350$.
Two Orbbec Gemini 335 cameras stream at 30 FPS, one mounted on the wrist and one fixed on a tripod beside the table as the third person agent view camera.
The image based policies observe the RGB images of both cameras, and the point cloud baselines build their point clouds from the aligned depth of the third person camera.
Policy inference runs on a separate machine with an NVIDIA RTX Pro 6000 Blackwell GPU, an AMD Ryzen Threadripper 7970X CPU, and 256 GB of memory.
\paragraph{Lighting Conditions.}
All demonstrations and evaluation trials are conducted at night, so daylight contributes to neither lighting condition.
Under nominal lighting, the scene is lit by the overhead laboratory lights.
The reduced lighting condition replaces them with two studio lights placed on either side of the workspace, which dim the illumination and make it directional while each task setup is held fixed.
Because the two cameras view the scene from different positions, the same change in lighting lowers their scene average luminance by different amounts.
\paragraph{Tasks.}
The three tasks are chosen to contrast manipulation that does not depend on color with manipulation in which color is the task signal.
Fruit is a simple pick and place task without a color cue, requiring the robot to pick up a cherry and place it in a bowl.
Bean uses color as the signal for sorting, requiring the robot to move the red beanbag into a tray with a blue beanbag on the table as a distractor.
Eraser uses color as the signal for long-horizon, contact-rich tool use, requiring the robot to grasp an eraser and wipe the red marking from a whiteboard that also carries a green marking.
\paragraph{Demonstrations and Evaluation.}
For each task, a human operator collects 40 demonstrations under nominal lighting by teleoperating the arm with a Meta Quest controller, and the demonstrations are recorded at 20 Hz.
Every policy is trained on these demonstrations and evaluated without adaptation over 10 trials under each lighting condition.
Example rollouts from the real world evaluation are included in the supplementary video.
\section{Experiments}
This section reports additional results and analyses for the simulated evaluation and real world evaluations of the main paper.
For LIBERO-Plus, success rates are reported separately for the variants that change light intensity, direction, specular strength, or shadow casting.
Keypoint drift between matched clean and perturbed renders of the same scene state isolates the change in visual conditioning that lighting alone produces.
The orbit distance of Appendix~\ref{apx:quant_invariance} is compared with keypoint drift on luminance shifts constructed at and between the group elements and on the perturbations of luminance shifted LIBERO and LIBERO-Plus.
The demonstration scaling study is described through the sampling of its demonstration subsets, the training recipe applied to them, and the initial states from which each setting is evaluated.
The failure modes of the real world evaluation are characterized through the behavior of the point cloud and image based baselines under reduced lighting.
\subsection{LIBERO-Plus}\label{appendix:exp_libero_plus}
\paragraph{Keypoint Drift.}
\begin{table}[t]
    \centering
    \caption{\textbf{LIBERO-Plus Lighting Keypoint Drift.} We report the keypoint drift of single-task policies between matched clean and LIBERO-Plus lighting renders. CARROT drifts the least in every suite.}
    \label{tab:libero_plus_keypoint_error}
    \small
    \renewcommand{\arraystretch}{0.9}
    \begin{tabular}{lcccc}
        \toprule
        Network & Goal & Object & Spatial & Overall\\
        \midrule
        Diffusion Policy & 0.1593 & 0.2335 & 0.2532 & 0.2154\\
        RetinexNet & 0.1232 & \underline{\smash{0.0912}} & 0.1609 & 0.1198\\
        Zero-DCE & \underline{\smash{0.1029}} & 0.1011 & \underline{\smash{0.1154}} & \underline{\smash{0.1065}}\\
        Stem-OB & 0.1296 & 0.1953 & 0.1699 & 0.1649\\
        \midrule
        CARROT & \textbf{0.0863} & \textbf{0.0790} & \textbf{0.1046} & \textbf{0.0900} \\
        \bottomrule
    \end{tabular}
\end{table}
\begin{table}[b]
    \centering
    \caption{\textbf{LIBERO-Plus Lighting Variable Intensity Success Rate.} Single-task success on the Goal, Object, and Spatial suites under LIBERO-Plus lighting variants that change the light intensity.}
    \label{tab:libero_plus_lighting_intensity_pooled}
    \small
    \setlength{\tabcolsep}{2pt}
    \renewcommand{\arraystretch}{0.9}
    \begin{tabular}{lcccc}
        \toprule
        Network & Goal & Obj. & Spa. & Avg.\\
        \midrule
        Diffusion Policy & $47.4{\scriptstyle\pm2.6}$ & $47.0{\scriptstyle\pm1.6}$ & $25.5{\scriptstyle\pm7.9}$ & $39.9{\scriptstyle\pm3.8}$\\
        DP-Jitter & $58.3{\scriptstyle\pm11.1}$ & $42.9{\scriptstyle\pm10.7}$ & $49.7{\scriptstyle\pm5.3}$ & $50.1{\scriptstyle\pm6.4}$\\
        DP-Planck & $\underline{\smash{64.2{\scriptstyle\pm6.0}}}$ & $41.1{\scriptstyle\pm7.2}$ & $36.8{\scriptstyle\pm4.3}$ & $47.0{\scriptstyle\pm5.4}$\\
        DP-DeepAug & $57.3{\scriptstyle\pm7.1}$ & $35.9{\scriptstyle\pm14.3}$ & $47.5{\scriptstyle\pm4.9}$ & $46.7{\scriptstyle\pm5.5}$\\
        RetinexNet & $32.5{\scriptstyle\pm3.6}$ & $41.3{\scriptstyle\pm6.4}$ & $41.2{\scriptstyle\pm7.6}$ & $38.4{\scriptstyle\pm3.8}$\\
        Zero-DCE & $41.8{\scriptstyle\pm4.7}$ & $\underline{\smash{58.6{\scriptstyle\pm10.8}}}$ & $\underline{\smash{56.6{\scriptstyle\pm12.1}}}$ & $\underline{\smash{52.5{\scriptstyle\pm8.4}}}$\\
        Stem-OB & $63.0{\scriptstyle\pm4.9}$ & $33.3{\scriptstyle\pm2.9}$ & $38.1{\scriptstyle\pm9.0}$ & $44.5{\scriptstyle\pm1.4}$\\
        \midrule
        CARROT & $\mathbf{71.4{\scriptstyle\pm8.8}}$ & $\mathbf{67.1{\scriptstyle\pm6.3}}$ & $\mathbf{71.7{\scriptstyle\pm6.5}}$ & $\mathbf{70.1{\scriptstyle\pm4.0}}$\\
        \bottomrule
    \end{tabular}
\end{table}
We measure how far the visual conditioning of each encoder moves when only the lighting changes.
For each of the 868 LIBERO-Plus lighting variants of the 30 tasks, we render the first policy observation twice, once in the perturbed environment and once in the matched clean environment that shares its initial state.
Both renders are taken on both cameras at the policy-native $128\times128$ resolution, after the initial state is set and the scene settles over five zero-action steps.
The two frames share a scene state, so any movement in the encoder's output comes from the lighting alone.
We run each frame unchanged through the evaluation pipeline of its method, namely the usual normalization and $116\times116$ center crop.
For RetinexNet and Zero-DCE, restoration precedes this pipeline and is applied identically to the clean and the perturbed frame.
From each encoding we read the 32 spatial-softmax keypoints, the softmax $\left(x,y\right)$ locations taken before any trailing linear layer, so we compare every method on the same quantity.
Drift for a pair is the mean Euclidean displacement of channel-matched keypoints,
\begin{equation}
    \mathrm{drift}\left(\nu,\mathrm{cam}\right) = \frac{1}{32}\sum_{i=1}^{32}\left\lVert k_{i}^{\mathrm{perturbed}}-k_{i}^{\mathrm{clean}}\right\rVert^{2},
\end{equation}
in the $\left[-1,1\right]$ crop frame, averaged over the two cameras.
We report task means, averaging variants within a task and then across the tasks of each suite, with the overall value taken across all 30 tasks.
A lighting-invariant encoder would score zero.
These perturbations are naturalistic, changing light intensity, direction, specular strength, and shadow casting, and they sit off the luminance orbit, so no method is invariant to them, CARROT included.
Table~\ref{tab:libero_plus_keypoint_error} shows that CARROT nonetheless drifts the least in every suite.
\paragraph{Success by Lighting Factor.}
\begin{table}[t]
    \centering
    \caption{\textbf{LIBERO-Plus Lighting Variable Direction Success Rate.} Single-task success on the Goal, Object, and Spatial suites under LIBERO-Plus lighting variants that change the light direction.}
    \label{tab:libero_plus_lighting_direction_pooled}
    \small
    \setlength{\tabcolsep}{2pt}
    \renewcommand{\arraystretch}{0.9}
    \begin{tabular}{lcccc}
        \toprule
        Network & Goal & Obj. & Spa. & Avg.\\
        \midrule
        Diffusion Policy & $46.5{\scriptstyle\pm1.9}$ & $49.2{\scriptstyle\pm2.0}$ & $26.0{\scriptstyle\pm8.3}$ & $40.6{\scriptstyle\pm3.9}$\\
        DP-Jitter & $57.2{\scriptstyle\pm11.9}$ & $43.8{\scriptstyle\pm9.7}$ & $49.4{\scriptstyle\pm5.9}$ & $50.0{\scriptstyle\pm6.1}$\\
        DP-Planck & $\underline{\smash{63.6{\scriptstyle\pm6.8}}}$ & $42.0{\scriptstyle\pm6.5}$ & $38.2{\scriptstyle\pm5.1}$ & $47.8{\scriptstyle\pm5.7}$\\
        DP-DeepAug & $55.7{\scriptstyle\pm7.5}$ & $36.3{\scriptstyle\pm14.5}$ & $49.7{\scriptstyle\pm4.2}$ & $47.0{\scriptstyle\pm6.0}$\\
        RetinexNet & $32.2{\scriptstyle\pm4.3}$ & $42.3{\scriptstyle\pm6.0}$ & $39.1{\scriptstyle\pm7.4}$ & $37.9{\scriptstyle\pm3.5}$\\
        Zero-DCE & $41.5{\scriptstyle\pm5.1}$ & $\underline{\smash{60.4{\scriptstyle\pm10.8}}}$ & $\underline{\smash{56.7{\scriptstyle\pm13.5}}}$ & $\underline{\smash{53.0{\scriptstyle\pm8.8}}}$\\
        Stem-OB & $62.9{\scriptstyle\pm5.4}$ & $34.6{\scriptstyle\pm3.2}$ & $39.3{\scriptstyle\pm9.7}$ & $45.4{\scriptstyle\pm1.4}$\\
        \midrule
        CARROT & $\mathbf{71.5{\scriptstyle\pm8.2}}$ & $\mathbf{66.2{\scriptstyle\pm6.5}}$ & $\mathbf{71.8{\scriptstyle\pm7.8}}$ & $\mathbf{69.8{\scriptstyle\pm3.8}}$\\
        \bottomrule
    \end{tabular}
\end{table}
\begin{table}[b]
    \centering
    \caption{\textbf{LIBERO-Plus Lighting Variable Specular Success Rate.} Single-task success on the Goal, Object, and Spatial suites under LIBERO-Plus lighting variants that change the specular term.}
    \label{tab:libero_plus_lighting_specular_pooled}
    \small
    \setlength{\tabcolsep}{2pt}
    \renewcommand{\arraystretch}{0.9}
    \begin{tabular}{lcccc}
        \toprule
        Network & Goal & Obj. & Spa. & Avg.\\
        \midrule
        Diffusion Policy & $48.1{\scriptstyle\pm1.7}$ & $46.8{\scriptstyle\pm1.4}$ & $25.2{\scriptstyle\pm6.7}$ & $40.1{\scriptstyle\pm2.9}$\\
        DP-Jitter & $59.1{\scriptstyle\pm11.3}$ & $42.6{\scriptstyle\pm11.4}$ & $50.8{\scriptstyle\pm8.3}$ & $50.8{\scriptstyle\pm7.0}$\\
        DP-Planck & $\underline{\smash{64.5{\scriptstyle\pm5.8}}}$ & $41.0{\scriptstyle\pm7.2}$ & $37.8{\scriptstyle\pm5.6}$ & $47.7{\scriptstyle\pm5.8}$\\
        DP-DeepAug & $58.8{\scriptstyle\pm6.9}$ & $35.1{\scriptstyle\pm13.1}$ & $49.2{\scriptstyle\pm4.7}$ & $47.6{\scriptstyle\pm5.3}$\\
        RetinexNet & $33.3{\scriptstyle\pm3.9}$ & $40.7{\scriptstyle\pm6.2}$ & $41.8{\scriptstyle\pm7.7}$ & $38.6{\scriptstyle\pm3.4}$\\
        Zero-DCE & $42.4{\scriptstyle\pm5.1}$ & $\underline{\smash{57.4{\scriptstyle\pm11.0}}}$ & $\underline{\smash{57.6{\scriptstyle\pm11.2}}}$ & $\underline{\smash{52.5{\scriptstyle\pm8.1}}}$\\
        Stem-OB & $63.9{\scriptstyle\pm5.1}$ & $32.3{\scriptstyle\pm3.8}$ & $38.9{\scriptstyle\pm8.5}$ & $44.9{\scriptstyle\pm1.1}$\\
        \midrule
        CARROT & $\mathbf{72.0{\scriptstyle\pm9.0}}$ & $\mathbf{64.8{\scriptstyle\pm7.7}}$ & $\mathbf{72.2{\scriptstyle\pm5.1}}$ & $\mathbf{69.6{\scriptstyle\pm4.8}}$\\
        \bottomrule
    \end{tabular}
\end{table}
Tables~\ref{tab:libero_plus_lighting_intensity_pooled} to~\ref{tab:libero_plus_lighting_shadow_pooled} break down the LIBERO-Plus lighting results of Table~\ref{tab:libero_plus_lighting} by lighting factor, namely light intensity, direction, specular strength, and shadow casting.
Intensity variants change image luminance almost uniformly, resembling a global luminance shift, whereas changes in direction, specular strength, and shadow casting vary the illumination across the scene in ways that no global luminance shift represents.
CARROT achieves the highest success in every suite for all four factors, so its advantage extends from near uniform changes to spatially varying ones.
\begin{table}[t]
    \centering
    \caption{\textbf{LIBERO-Plus Lighting Variable Shadow Success Rate.} Single-task success on the Goal, Object, and Spatial suites under LIBERO-Plus lighting variants that add shadow casting.}
    \label{tab:libero_plus_lighting_shadow_pooled}
    \small
    \setlength{\tabcolsep}{2pt}
    \renewcommand{\arraystretch}{0.9}
    \begin{tabular}{lcccc}
        \toprule
        Network & Goal & Obj. & Spa. & Avg.\\
        \midrule
        Diffusion Policy & $47.0{\scriptstyle\pm5.3}$ & $45.3{\scriptstyle\pm2.1}$ & $26.9{\scriptstyle\pm7.6}$ & $39.9{\scriptstyle\pm4.7}$\\
        DP-Jitter & $61.6{\scriptstyle\pm9.8}$ & $41.4{\scriptstyle\pm10.5}$ & $47.5{\scriptstyle\pm6.1}$ & $50.0{\scriptstyle\pm6.0}$\\
        DP-Planck & $\underline{\smash{64.8{\scriptstyle\pm4.4}}}$ & $37.6{\scriptstyle\pm8.7}$ & $37.6{\scriptstyle\pm6.1}$ & $46.6{\scriptstyle\pm5.9}$\\
        DP-DeepAug & $59.3{\scriptstyle\pm7.6}$ & $34.7{\scriptstyle\pm12.8}$ & $45.4{\scriptstyle\pm5.2}$ & $46.2{\scriptstyle\pm5.5}$\\
        RetinexNet & $33.5{\scriptstyle\pm2.5}$ & $39.0{\scriptstyle\pm4.5}$ & $39.4{\scriptstyle\pm7.9}$ & $37.3{\scriptstyle\pm3.6}$\\
        Zero-DCE & $42.8{\scriptstyle\pm3.5}$ & $\underline{\smash{57.8{\scriptstyle\pm9.7}}}$ & $\underline{\smash{57.4{\scriptstyle\pm14.1}}}$ & $\underline{\smash{52.8{\scriptstyle\pm8.4}}}$\\
        Stem-OB & $62.5{\scriptstyle\pm4.1}$ & $32.0{\scriptstyle\pm3.3}$ & $36.1{\scriptstyle\pm8.9}$ & $43.3{\scriptstyle\pm0.9}$\\
        \midrule
        CARROT & $\mathbf{68.7{\scriptstyle\pm9.4}}$ & $\mathbf{67.6{\scriptstyle\pm7.0}}$ & $\mathbf{71.0{\scriptstyle\pm7.7}}$ & $\mathbf{69.1{\scriptstyle\pm4.5}}$\\
        \bottomrule
    \end{tabular}
\end{table}
\subsection{Orbit-Distance Analysis}\label{apx:orbit_distance}
\begin{figure}[h]
    \centering
    \includegraphics[width=\columnwidth]{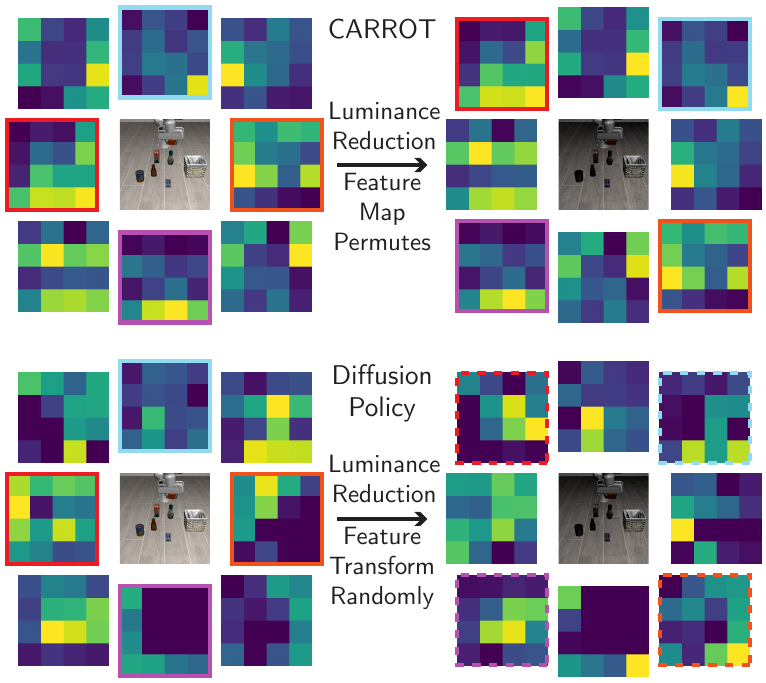}
    \caption{\textbf{Feature Map Permutation.} We show feature maps of CARROT and Diffusion Policy before and after a modeled luminance reduction. Matching colors marks corresponding maps and dashed outlines marks where a permutation would place each Diffusion Policy map. The feature maps of CARROT permute exactly across the luminance orbit, while the feature maps of Diffusion Policy change without structure.}
    \label{fig:feature_map_permutation}
\end{figure}

\begin{figure}[b]
    \centering
    \includegraphics[width=\columnwidth]{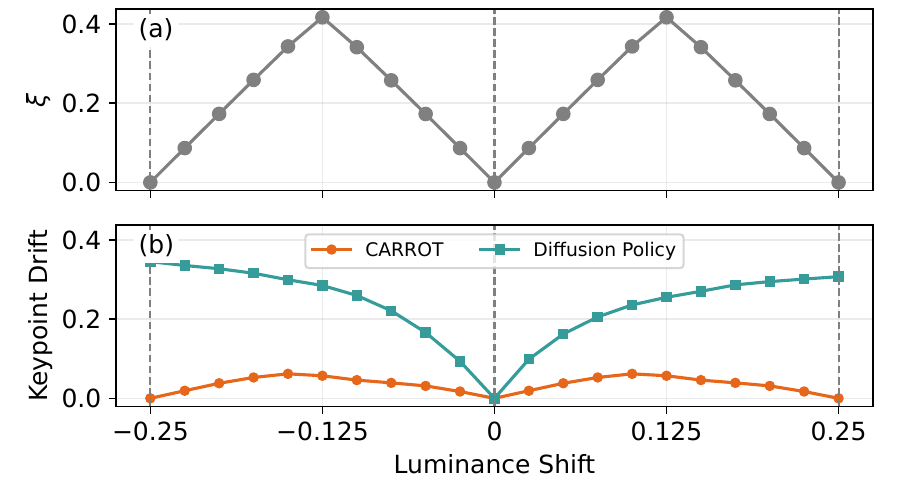}
    \caption{\textbf{Shifts Between Group Elements.} We report the \textbf{(a)} relative orbit distance $\xi$ and the \textbf{(b)} keypoint drift of CARROT and diffusion policy under uniform luminance shifts without clipping. The orbit distance and CARROT drift vanish at the group elements.}
    \label{fig:CARROT_Invariance}
\end{figure}

This section measures the orbit distance of Appendix~\ref{apx:quant_invariance} on controlled luminance shifts constructed for this analysis and on the perturbations of the evaluated benchmarks, and compares it with the keypoint drift of CARROT and Diffusion Policy.
\paragraph{Exact Invariance.}
We first verify exact invariance on frames whose luminance is compressed to $[0.25,0.75]$, so that shifts of $\pm0.25$ are modeled luminance shifts.
Figure~\ref{fig:feature_map_permutation} shows that a modeled luminance reduction cyclically permutes the feature maps of CARROT across the luminance orbit, with each map reappearing unchanged one group element away, whereas the feature maps of Diffusion Policy change without any such structure.
As shown in Table~\ref{tab:exact_invariance}, the closest element matches the applied shift for every pair, and because orbit pooling absorbs this permutation, CARROT keypoints remain fixed to numerical precision while Diffusion Policy keypoints move.
When a block of pixels is pushed across the fold onto the upper branch, the entire residual orbit distance lies on that block.
\begin{table}[b]
    \centering
    \caption{\textbf{Exact Invariance.} We report the relative orbit distance, the recovered group element, and keypoint drift under modeled shifts of $\pm0.25$ on luminance-compressed frames. CARROT is invariant up to numerical precision.}
    \label{tab:exact_invariance}
    \small
    \renewcommand{\arraystretch}{0.9}
    \begin{tabular}{lc}
        \toprule
        Quantity & Value\\
        \midrule
        Relative orbit distance $\xi$, worst pair & $1.3\times10^{-7}$\\
        CARROT keypoint drift, worst pair & $5.1\times10^{-5}$\\
        \midrule
        Diffusion policy keypoint drift, median & $0.23$\\
        \bottomrule
    \end{tabular}
\end{table}
\paragraph{Shifts Between Group Elements.} 
Figure~\ref{fig:CARROT_Invariance} sweeps uniform shifts between the group elements on the same compressed frames. 
The relative orbit distance vanishes at shifts of $0$ and $\pm0.25$ and grows linearly with the offset from the nearest of these, peaking midway at $\pm0.125$, consistent with the linear dependence on this offset in Appendix~\ref{apx:quant_invariance}. 
CARROT drift likewise vanishes at the group elements and is largest between them, while diffusion policy drift increases monotonically with the magnitude of the shift.
\paragraph{Benchmark Perturbations.} 
Table~\ref{tab:orbit_distance_benchmark} reports the orbit distance and keypoint drift under the two evaluated benchmarks.
Under the luminance-shifted benchmark, the closest orbit element is $l_{7}$ for every frame, and the residual orbit distance lies entirely on pixels clipped to zero, with its size tracking the share of such pixels.
The LIBERO-Plus lighting changes lie considerably further from the orbit, as expected for spatially varying and chromatic illumination that no global luminance shift reproduces.
CARROT nonetheless drifts less than diffusion policy on both benchmarks and both cameras, indicating that the structured representation remains useful beyond the modeled action.
\begin{table}[t]
    \centering
    \caption{\textbf{Orbit Distance of Benchmark Perturbations.} We report the median unabsorbed fraction $\eta$, the median relative orbit distance $\xi$, and the median keypoint drift (mean $\pm$ std over three training seeds) for each camera. The luminance shift lies close to the orbit, while the LIBERO-Plus lighting changes lie far from it.}
    \label{tab:orbit_distance_benchmark}
    \small
    \setlength{\tabcolsep}{4pt}
    \renewcommand{\arraystretch}{0.9}
    \begin{tabular}{lcccc}
        \toprule
        \multirow{2}{*}{Camera} & \multirow{2}{*}{$\eta$} & \multirow{2}{*}{$\xi$} & \multicolumn{2}{c}{Keypoint Drift}\\
        \cmidrule(lr){4-5}
        & & & CARROT & DP\\
        \midrule
        \multicolumn{5}{l}{Luminance-Shifted LIBERO}\\
        \addlinespace[3pt]
        Agentview & $0.25$ & $0.19$ & $\mathbf{0.048{\scriptstyle\pm0.004}}$ & $0.101{\scriptstyle\pm0.001}$\\
        Wrist & $0.24$ & $0.19$ & $\mathbf{0.082{\scriptstyle\pm0.017}}$ & $0.327{\scriptstyle\pm0.015}$\\
        \midrule
        \multicolumn{5}{l}{LIBERO-Plus Lighting}\\
        \addlinespace[3pt]
        Agentview & $0.72$ & $0.53$ & $\mathbf{0.076{\scriptstyle\pm0.006}}$ & $0.118{\scriptstyle\pm0.014}$\\
        Wrist & $0.55$ & $0.40$ & $\mathbf{0.103{\scriptstyle\pm0.020}}$ & $0.369{\scriptstyle\pm0.014}$\\
        \bottomrule
    \end{tabular}
\end{table}
\subsection{Demonstration Scaling}\label{apx:demo_scaling}
The policies in Figure~\ref{fig:CARROT_sample} are trained on subsets of the Spatial suite demonstrations and evaluated on unperturbed LIBERO and under LIBERO-Plus lighting over three training seeds.
\paragraph{Demonstration Subsets and Training.} 
For each task, the 10 and 20 demonstration subsets are drawn at random from the 50 demonstrations with a fixed seed, so that every training seed and every method uses the same subset. 
Stem-OB trains on the inverted counterparts of the same demonstrations. 
The 50 demonstration policies are those of the main experiments. 
All policies train on every demonstration in their subset under the recipe of the main experiments, with the number of epochs held fixed so that the number of gradient steps grows in proportion to the number of demonstrations.
Every policy is evaluated at its final checkpoint with EMA weights.
\paragraph{Evaluation.} 
The two settings start from different initial states. On unperturbed LIBERO, each policy is rolled out from the 50 canonical test initial states of each task.
Under LIBERO-Plus lighting, each policy starts from the robot states of the training demonstrations, with the lighting perturbations applied as in Table~\ref{tab:libero_plus_lighting}.
\subsection{Failure Modes Under Reduced Lighting}\label{apx:failure_modes}
\begin{figure*}[tp]
    \centering
    \includegraphics[width=\textwidth,height=0.9\textheight,keepaspectratio]{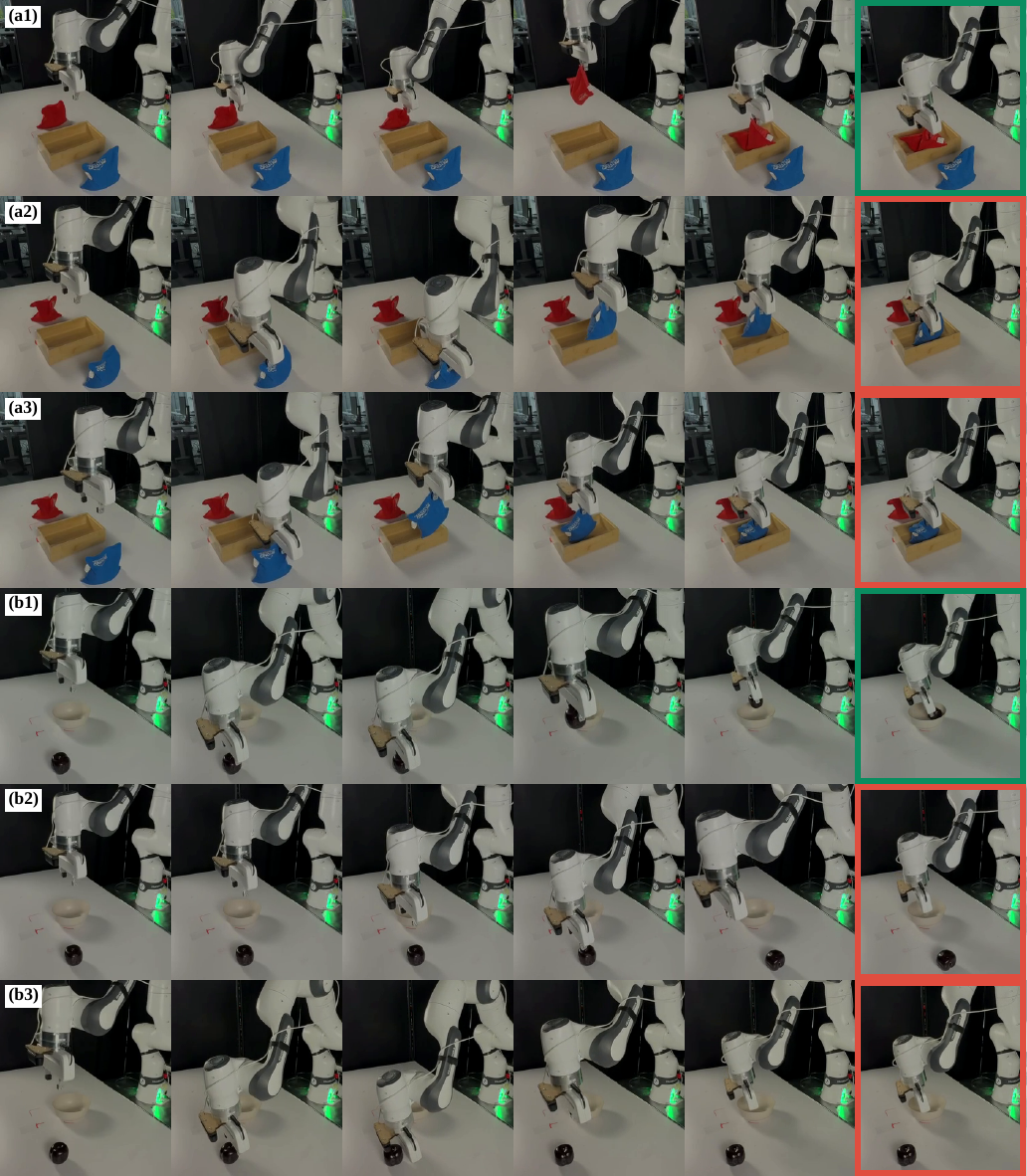}
    \caption{\textbf{Failure Modes under Reduced Lighting.} Keyframes from the Bean rollout with the red beanbag on the far side of the tray for \textbf{(a1)} CARROT, \textbf{(a2)} DP3-RGB, and \textbf{(a3)} DP3, and from the Fruit rollout for \textbf{(b1)} CARROT, \textbf{(b2)} Diffusion Policy, and \textbf{(b3)} Zero-DCE. Time runs left to right, and the final frame is outlined in green for success and red for failure. DP3-RGB and DP3 place the blue beanbag in the tray, while Diffusion Policy and Zero-DCE lift without the cherry and end over the bowl with an empty gripper.}
    \label{fig:failure_filmstrip}
\end{figure*}
This section characterizes the baseline failures in the real world evaluation under reduced lighting, drawing on the rollouts in the supplementary video and the keyframes in Figure~\ref{fig:failure_filmstrip}.
The point cloud baselines (DP3 and DP3-RGB) are studied on the Bean and Eraser tasks, in which color is the only property that distinguishes the target from a distractor.
The image based baselines (Diffusion Policy, Zero-DCE, and Stem-OB) are studied across all three tasks, with attention to whether the gripper secures the object once it reaches it.
\paragraph{Point Cloud Baselines.}
DP3 and DP3-RGB complete the Fruit task, which involves a single object, and on the color-sensitive Bean and Eraser tasks most of their failures follow a clean grasp of the wrong object.
With the red beanbag on the far side of the tray, DP3 and DP3-RGB each grasp the blue beanbag and place it in the tray (Figure~\ref{fig:failure_filmstrip}, a2 and a3).
With the red marking on the far side of the whiteboard, both pick up the eraser and bring it to the green marking while the red marking stays on the board.

For DP3 this outcome follows from its input.
Both tasks specify the target by color, which a colorless point cloud does not carry, and the whiteboard markings add no measurable depth, so DP3 cannot perceive either marking.
DP3-RGB succeeds on both color-sensitive tasks under nominal lighting (Table~\ref{tab:realworld_light}) yet makes the same choices as DP3 in these rollouts, so the color attached to its point cloud no longer separates the target from the distractor once the lighting is reduced.
Discarding appearance, as DP3 does, makes a policy largely insensitive to illumination by construction, but it also removes the cue that defines the target in these tasks.
CARROT keeps color in its input and handles the luminance change through the equivariance of its encoder, and it selects the red beanbag and the red marking in the same rollouts.
\paragraph{Image Based Baselines.}
The image based baselines generally move toward the correct object but fail to secure it.
In the Fruit task, Diffusion Policy and Zero-DCE descend onto the cherry and lift without it, and all three baselines end the rollout with an empty gripper over the bowl while the cherry remains on the table (Figure~\ref{fig:failure_filmstrip}, b2 and b3).
With the red marking on the far side of the whiteboard, all three approach the eraser without lifting it and then move over the red marking while the eraser stays on its box.
With the red beanbag on the far side of the tray, Diffusion Policy and Stem-OB reach the red beanbag without lifting it, and Zero-DCE hovers above the tray for the whole rollout.

The arm motion in these rollouts stays coherent and is directed at the correct object, while the grasp itself fails.
Neither the enhancement in Zero-DCE nor the inversion in Stem-OB removes this pattern.
The two groups of baselines thus fail in complementary ways.
The point cloud baselines secure the object without the color cue that identifies it, and the image based baselines retain enough appearance to find the target but not enough to secure it.
CARROT completes both stages in every rollout shown, which is consistent with an encoder whose features transform predictably under the luminance change rather than degrading with it.
\section{Training and Evaluation}\label{apx:training}
\begin{table}[t]
    \centering
    \caption{\textbf{Hyperparameters.} We report the configuration of the image based policies in simulation and on the real robot. Single task CARROT and Diffusion Policy differ only in the encoder.}
    \label{tab:hyperparameters}
    \small
    \setlength{\tabcolsep}{4pt}
    \begin{tabular}{lcc}
        \toprule
        Hyperparameter & Simulation & Real World\\
        \midrule
        \multicolumn{3}{l}{\textbf{Policy}}\\
        \addlinespace[3pt]
        Observation horizon & $2$ & $2$\\
        Prediction horizon & $16$ & $16$\\
        Action horizon & $8$ & $8$\\
        U-Net channels & $512/1024/2048$ & $512/1024/2048$\\
        Kernel size & $5$ & $5$\\
        Step embedding & $128$ & $128$\\
        Keypoints per camera & $32$ & $32$\\
        Group order $R$ & $8$ & $8$\\
        \midrule
        \multicolumn{3}{l}{\textbf{Observation}}\\
        \addlinespace[3pt]
        Image size & $128\times128$ & $320\times240$\\
        Crop size & $116\times116$ & $288\times216$\\
        \midrule
        \multicolumn{3}{l}{\textbf{Diffusion}}\\
        \addlinespace[3pt]
        Diffusion steps & $100$ & $100$\\
        Noise schedule & cosine & cosine\\
        Prediction target & noise & noise\\
        Evaluation sampler & DDPM, $100$ steps & DDIM, $16$ steps\\
        \midrule
        \multicolumn{3}{l}{\textbf{Optimization}}\\
        \addlinespace[3pt]
        Optimizer & AdamW & AdamW\\
        Learning rate & $10^{-4}$ & $10^{-4}$\\
        Betas & $\left(0.95, 0.999\right)$ & $\left(0.95, 0.999\right)$\\
        Weight decay & $10^{-6}$ & $10^{-6}$\\
        Warmup steps & $500$ & $500$\\
        LR schedule & cosine & cosine\\
        Batch size & $64$ & $64$\\
        Epochs & $1000$ & $1500$\\
        EMA power & $0.75$ & $0.75$\\
        EMA maximum decay & $0.9999$ & $0.9999$\\
        \bottomrule
    \end{tabular}
\end{table}
This section specifies the configuration, training procedure, and evaluation protocol of the policies in the main paper, both in simulation and on the real robot.
The policy configuration, shared by every image based policy, follows Diffusion Policy~\citep{chi2025diffusion} and is described through the denoiser, the noise schedule, the observation and action horizons, and the normalization of observations and actions.
The training procedure is specified through the optimizer, the learning rate schedule, the training budget, and the checkpoint and weight averaging used at evaluation.
Table~\ref{tab:hyperparameters} lists the corresponding values for simulation and for the real robot.
The evaluation protocol states how simulated rollouts are initialized, terminated, and aggregated into the reported success rates.
For the real robot, it states how each predicted action sequence is executed and how the inference latency of Table~\ref{tab:latency} is timed.
\paragraph{Policy Configuration.}
Diffusion Policy, Stem-OB, and CARROT share the configuration of~\citet{chi2025diffusion}, listed in Table~\ref{tab:hyperparameters}.
The single task configurations of CARROT and Diffusion Policy differ only in the encoder, and Stem-OB differs from Diffusion Policy only in the frames on which it is trained.
Every encoder is trained from scratch, as in Diffusion Policy.
The denoiser is a one dimensional convolutional U-Net conditioned through FiLM~\citep{perez2018film} on the diffusion step embedding and on the visual and proprioceptive features of both observation steps.
For language conditioning, the 768 dimensional BERT~\citep{devlin2019bert} task embedding is projected to 128 dimensions by a two layer MLP and appended to this conditioning.
DP3 and DP3-RGB follow the reference implementation of~\citet{ze20243d} and share the horizons, proprioceptive state, and action space of the image based policies.
Their denoisers are trained to predict the clean action sequence rather than the added noise.
\paragraph{Observations and Actions.}
In simulation, both cameras render $128\times128$ frames, which are cropped to $116\times116$.
The nine dimensional proprioceptive state contains the seven joint angles and the two gripper joint positions, and each action is the seven dimensional delta end effector command of LIBERO, with three translational, three rotational, and one gripper component.
On the real robot, frames from both cameras are resized from $640\times480$ to $320\times240$ pixels and cropped to $288\times216$.
The proprioceptive state concatenates the seven joint angles, the end effector position and orientation, and the gripper width, and each action concatenates the absolute target end effector position and orientation with a binary gripper command.
Both orientations are expressed in the continuous 6D rotation representation~\citep{zhou2019continuity}, which encodes a rotation by two vectors of its rotation matrix and, unlike Euler angles and quaternions, has no discontinuities.
The real world state therefore has 17 dimensions and the action has 10.
In both settings, crops are taken at a random position during training and at the center during evaluation.
Images and states are rescaled to $\left[-1,1\right]$, with the range of each state dimension taken from the demonstrations.
Real world actions are rescaled in the same way, whereas the simulated delta actions already lie in $\left[-1,1\right]$.
\paragraph{Training.}
Each policy is trained on a single task, or on all ten tasks of a suite in the language conditioned setting, with the optimizer, learning rate schedule, and weight averaging of Table~\ref{tab:hyperparameters}, for a fixed number of epochs.
Random cropping is the only data augmentation of Diffusion Policy and CARROT.
Every policy is evaluated with the exponential moving average of its weights at the last checkpoint, without selection by validation loss or rollout success.
In simulation, each policy is trained on a single NVIDIA RTX Pro 6000 Blackwell GPU with the three training seeds 42, 72, and 172, except that the language conditioned augmentation baselines use only seed 42.
The language conditioned CARROT policies are instead trained with data parallelism on four GPUs with 64 samples per GPU and a learning rate of $2\times10^{-4}$, for the same number of epochs as the language conditioned baselines.
On the real robot, each image based policy is trained once, with seed 42, using distributed data parallelism on two NVIDIA RTX Pro 6000 Blackwell GPUs with 32 samples per GPU.
DP3 and DP3-RGB use the same seed, optimizer, learning rate schedule, weight averaging, and number of epochs with a batch size of 128.
\paragraph{Evaluation.}
\begin{figure}[t]
    \centering
    \includegraphics[width=\columnwidth]{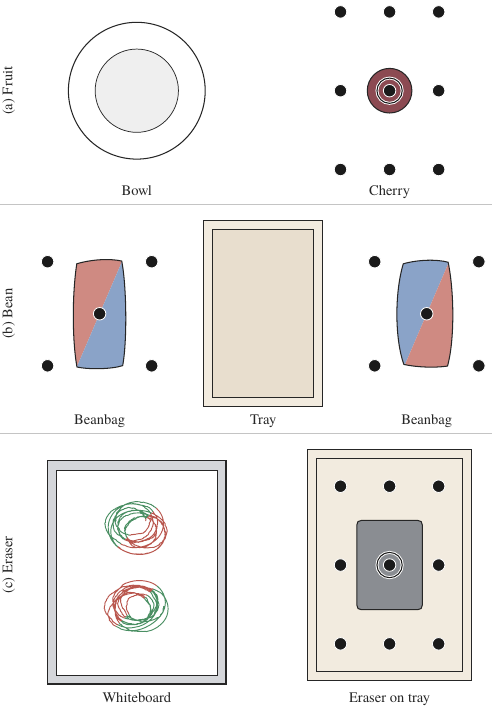}
    \caption{\textbf{Initial Configurations.} We show a bird's eye view of the initial configurations of each real world task, in which dots mark the starting centers of the moving objects and a double circle marks the central position, which appears twice among the ten configurations of the Fruit and Eraser tasks. Across configurations, the red and blue beanbags exchange sides of the tray and the red and green markings exchange places on the whiteboard, while the bowl, trays, and whiteboard remain fixed.}

    \label{fig:CARROT_InitialConfigs}
\end{figure}
In simulation, every policy is rolled out from the same 20 initial states per task, those of the first 20 demonstrations of the task, once on clean and once on luminance shifted observations, so that lighting is the only difference between the two conditions.
Each single task policy is also rolled out once in every LIBERO-Plus lighting variant of its task.
Policies sample with 100 DDPM~\citep{ho2020denoising} steps and execute eight actions of each prediction, and an episode ends at the first success or after 600 environment steps.
Each language conditioned policy is evaluated on clean observations and in the LIBERO-Plus lighting variants of all ten tasks of its suite.
Success rates are averaged over the tasks of each suite, and the tables report the mean and standard deviation over the three training seeds.
\paragraph{Real World Evaluation.}
Auto exposure, gain, and white balance are disabled on both cameras, which keep an exposure of 2600~$\mu$s, a gain of 0, and a white balance of 4600~K from demonstration collection through both evaluation conditions.
Each camera therefore applies the same response under both lighting conditions and does not compensate for the reduced light.
The luminance change $\Delta$L in Table~\ref{tab:realworld_dark} is the change in mean HSL lightness over the segmented table area relative to nominal lighting, measured once for each task from a single frame of each camera at $320\times240$ under each lighting condition, with the task setup held fixed.
The light sources are never moved, so this single measurement applies to every trial of the task.
Each task is evaluated from ten initial object configurations, shown in Figure~\ref{fig:CARROT_InitialConfigs}, which are shared by both lighting conditions.
Each configuration is run by CARROT and then by every baseline in sequence.
\begin{figure}[t]
    \centering
    \includegraphics[width=\columnwidth]{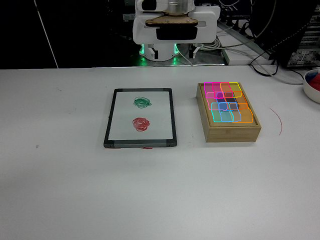}
    \caption{\textbf{Initial Configuration Overlay.} We show the reference outlines of the eraser in the initial configurations of the Eraser task, overlaid on a frame of the third person camera. Before each trial, the objects are repositioned until they coincide with the outline of the current configuration, so every policy starts from the same placement.}
    \label{fig:CARROT_InitialOverlay}
\end{figure}
Before every trial, the initial frame of the Diffusion Policy rollout from the same configuration is overlaid on the live camera view, as shown for the Eraser task in Figure~\ref{fig:CARROT_InitialOverlay}, and the objects are repositioned until they coincide with it, so that every policy starts from the same placement.
Every policy, including DP3 and DP3-RGB, samples with 16 DDIM~\citep{song2020denoising} steps over the training noise schedule, following the real world experiments of~\citet{chi2025diffusion}.
Of the 16 predicted actions, the first corresponds to the preceding observation step and is discarded, and the remaining actions are timestamped at intervals of 50 ms from the current observation and passed to a 20 Hz executor, which drops those already past and sends the rest to the Deoxys operational space controller.
A new prediction is requested every 400 ms, so eight actions of each prediction are executed, matching the action horizon.
A Fruit trial succeeds when the cherry rests in the bowl, a Bean trial when the red beanbag lies in the tray with the blue beanbag outside it, and an Eraser trial when the red marking is wiped off and the green marking remains.
Success is determined by an evaluator who is blind to the policy being run and has access to the stop button of the robot.
\paragraph{Inference Latency.}
Table~\ref{tab:latency} is measured at a batch size of one on a machine with a single NVIDIA RTX Pro 6000 Blackwell GPU, an AMD Ryzen Threadripper 7970X CPU, and 256 GB of memory, and reports the median of ten timed passes after warmup.
The forward time covers observation normalization, the encoder including the luminance lift of CARROT, all 16 denoising steps, and the transfer of the predicted actions to the CPU.
The total time adds image decoding and resizing, together with Zero-DCE restoration and the point cloud preprocessing of DP3 and DP3-RGB where they apply, and excludes network transfer.
\end{document}